\documentclass[11pt,a4paper]{article}

\usepackage{placeins}
\usepackage{booktabs} %
\usepackage{amsmath}
\usepackage{amssymb}  %
\usepackage{amsfonts} %
\usepackage{bm}       %
\usepackage{nicefrac} %
\usepackage{subcaption}  %
\usepackage{acronym}  %
\usepackage{multicol,multirow}
\usepackage{longtable}

\usepackage[nocustomfonts]{nxai}
\newcommand{\affmark}[1]{\raisebox{0.55ex}{\fontsize{7}{7}\selectfont #1}}

\nxaisetalogo{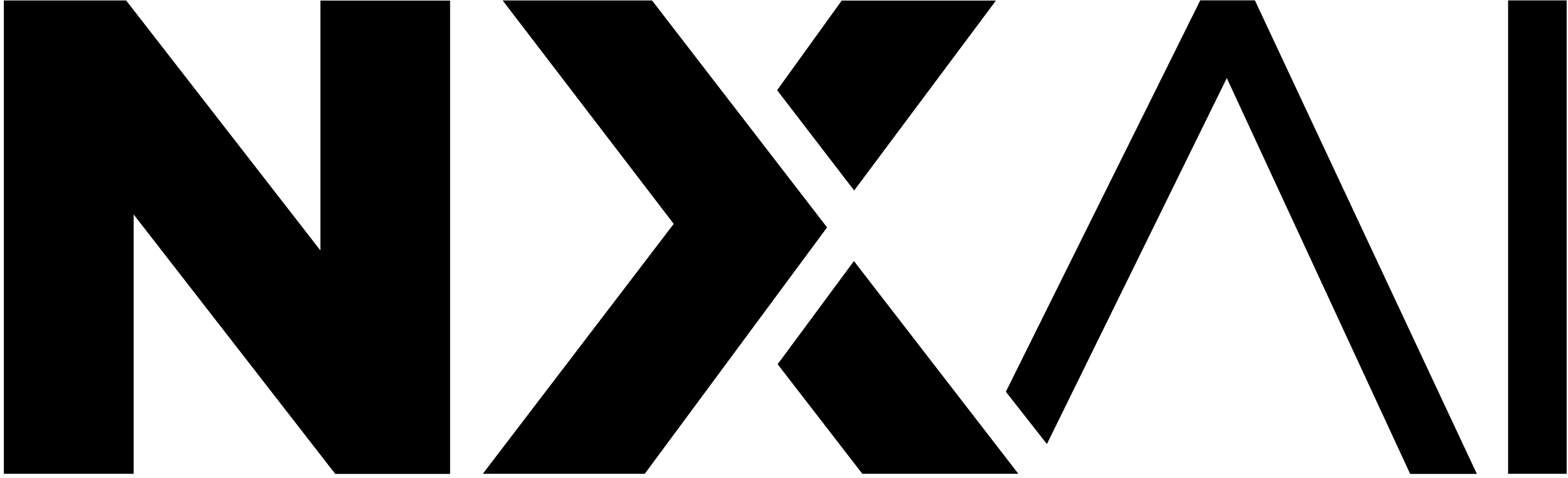} %
\nxaisetlogoheight{4mm} %
\nxaisetfooter{Corresponding authors: Anamaria-Roberta Hartl (hartl@ml.jku.at), Levente Zólyomi (levente.zolyomi@nx-ai.com)}

\title{On Subquadratic Architectures: From Applications to Principles}
\author{Anamaria-Roberta Hartl\affmark{*1} \and Levente Zólyomi\affmark{*1,2} \and David Stap\affmark{2} \and Pieter-Jan Hoedt\affmark{1} \and \\ Niklas Schmidinger\affmark{1,2} \and Lukas Hauzenberger\affmark{1,2} \and Sebastian Böck\affmark{2} \and Günter Klambauer\affmark{1,2} \and \\ Sepp Hochreiter\affmark{1,2}}
\nxaisetaffiliations{%
\affmark{1} Johannes Kepler University Linz \affmark{2} NXAI GmbH \affmark{*} Equal contribution.
}

\acrodef{tsfm}[TSFM]{Time Series Foundation Models}
\acrodef{mase}[MASE]{Mean Absolute Scaled Error}
\acrodef{crps}[CRPS]{Continuous Ranked Probability Score}
\acrodef{gru}[GRU]{Gated Recurrent Unit}

\ExplSyntaxOn
\cs_new_eq:NN \ToUppercase \text_uppercase:n
\cs_new_eq:NN \ToLowercase \text_lowercase:n
\ExplSyntaxOff
\newcommand\reals     {\mathbb{R}}
\renewcommand\vec[1]  {\bm{\ToLowercase{#1}}}
\newcommand\mat[1]    {\bm{\ToUppercase{#1}}}
\newcommand\T         {{\mkern-1.5mu\mathsf{T}}}
\newcommand\norm[1]   {\|#1\|}
\DeclareMathOperator\softmax   {softmax}
\DeclareMathOperator\softplus  {softplus}
\DeclareMathOperator\diag      {diag}

\definecolor{xlstm_accent}{RGB}{102,41,194}
\definecolor{mamba2_accent}{RGB}{216,6,30}
\definecolor{gdn_accent}{RGB}{254,122,7}

\nxaiabstract{%
Transformers dominate modern sequence modeling, but their quadratic attention incurs substantial computational cost.
Subquadratic architectures offer a scalable alternative.
However, it remains unclear which designs yield the most effective sequence models.
We compare three leading approaches: xLSTM, Mamba-2, and Gated DeltaNet.
We evaluate these models on tasks with complex dependencies: (1) code-model pre-training, (2) distillation of code models from large language models, and (3) pre-training of time-series foundation models.
Across these settings, xLSTM delivers the strongest overall performance.
To explain xLSTM's advantage, we present a unified formulation and analyze the underlying architectural mechanisms, focusing on state tracking and memory dynamics.
Our results show that xLSTM enables more flexible and stable memory correction via its gating scheme.
We corroborate these findings on controlled synthetic length-generalization tasks.
Overall, our findings indicate that xLSTM's gains on complex tasks stem from robust state tracking and accumulation.
}

\begin{document}
\maketitle

\section{Introduction}
\label{sec:introduction}

\paragraph{Subquadratic architectures as scalable alternatives to Transformers.}
Transformers~\citep{vaswani17attention} dominate modern sequence modeling and remain the default backbone for foundation models in language, code, and time series.
At the same time, their quadratic attention cost has motivated subquadratic alternatives based on recurrent, state-space, and linear-attention mechanisms.
Recent hybrid foundation models such as Samba~\citep{ren25samba}, Nemotron Nano~\citep{nvidia25nemotron3nano}, Kimi Linear~\citep{team25kimi}, and Olmo Hybrid~\citep{merrill26olmo_hybrid} replace many attention layers with subquadratic sequence operators.
This makes the choice of operator central to the design of modern hybrid language models.

\paragraph{Three leading architectures: xLSTM, Mamba-2 and Gated DeltaNet.}
Several subquadratic architectures have been suggested for a diverse range of tasks~\citep{fichtl25subquadratic}.
Out of those, xLSTM~\citep{beck24xlstm} has demonstrated competitive language modeling~\citep{beck25xlstm7b} and was shown to Pareto-dominate transformers \citep{beck26xlstm}.
Moreover, it serves as the backbone of TiRex, one of the best-performing time-series foundation models~\citep{auer25tirex}.
Mamba-2 \citep{dao24mamba2} appears as a core component in competitive hybrid language models \citep{lenz25jamba,glorioso24zamba,ren25samba,nvidia25nemotron3nano}.
Finally, Gated DeltaNet~\citep{yang24gdn} is used in competitive hybrid language models~\citep{team25kimi, merrill26olmo_hybrid}, and has been adopted in recent time-series work~\citep{moroshan25tempopfn}.
While prior work motivates all three architectures as relevant backbones, no head-to-head comparison exists.

\begin{figure}[h]
    \centering
    \begin{subfigure}[t]{0.49\textwidth}
        \centering
        \includegraphics[width=\linewidth]{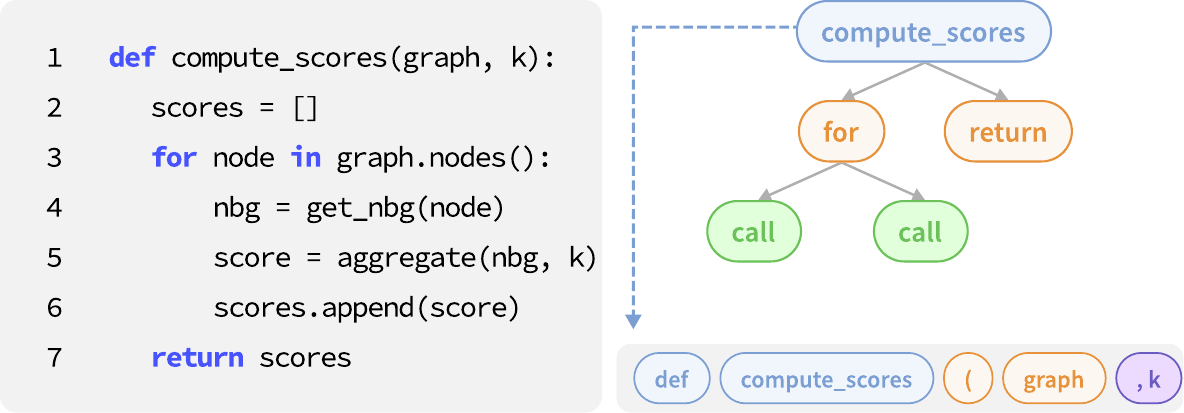}
        \caption{Code}
        \label{fig:complex-deps-code}
    \end{subfigure}
    \hfill
    \begin{subfigure}[t]{0.40\textwidth}
        \centering
        \includegraphics[width=\linewidth]{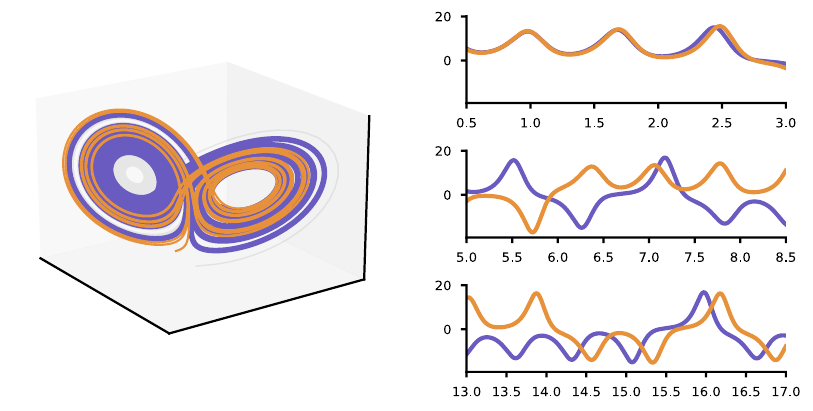}
        \caption{Time Series}
        \label{fig:complex-deps-ts}
    \end{subfigure}
    \caption{\textbf{Tasks with complex dependencies.} Code (a) carries dependencies in formal structure: syntax trees, call graphs, variable bindings. Time series (b) carries them in partially observed dynamics: trajectories of complex systems (here, a Lorenz attractor) whose future depends on unobserved states over history. Both are representative of complex dependencies where modeling requires tracking many interacting states across long contexts.}
    \label{fig:complex-deps}
\end{figure}

\paragraph{A comparison on complex data domains.}
So far, subquadratic backbones have mostly been compared on standard language modeling and commonsense reasoning benchmarks, where performance differences are small and architectures are hard to differentiate~\citep{yang24gla,mishra24lm_engine}.
In contrast, prior work has shown that architectural inductive biases diverge sharply on data with long-range, structured dependencies~\citep{deletang23chomsky,liu23exposing}.
We therefore evaluate the operators on \textit{complex data domains}: naturally-occurring data whose generating process imposes such structured dependencies, instantiated here by \textit{code} and \textit{time series} (see Figure \ref{fig:complex-deps}).
On the one hand, code combines language-like tokens with formal structure, including syntax, variable bindings, and scopes~\citep{siems26learning, merrill26olmo_hybrid}.
On the other hand, time series require models to infer and update complex dynamics from continuous-valued histories across heterogeneous domains~\citep{ansari24chronos,das24timesfm,auer25tirex}.
We test the sequence backbones under both from-scratch pre-training and Transformer-to-subquadratic distillation~\citep{schmidhuber1991neural, hinton15distilling, mercat24linearizing}, and additionally evaluate the models on code generation tasks.
Across these three settings, xLSTM-based backbones\footnote{The xLSTM family combines matrix-state linear-attention layers, denoted xLSTM$[1\!:\!0]$ or mLSTM, with recurrent layers, denoted xLSTM$[0\!:\!1]$ or sLSTM; xLSTM$[m\!:\!s]$ denotes the ratio of these two components.} consistently show favorable results, which raises the central question of the paper: \textit{which architectural design choices explain xLSTM's advantage on complex sequence tasks?}

\paragraph{A unified framework for xLSTM, Mamba-2, and Gated DeltaNet.}
We explain xLSTM's advantage by formulating xLSTM, Mamba-2, and Gated DeltaNet into a unified framework (Section~\ref{sec:framework}).
This formulation makes the architectures directly comparable at the level where they differ most: how they write, forget, overwrite, and read from state.
Our unified formulation identifies a hypothesis that the architectures should differ most on two primitive capabilities: accumulation and state tracking.
Our framework motivates to test this hypothesis on controlled synthetic length-generalization tasks (Section~\ref{sec:synthetic}).
Counting tasks isolate accumulation beyond the training length, while state-tracking tasks isolate ordered finite-state updates over sequences.
The results support this prediction, where xLSTM can solve both of these task families well beyond its training length.
Together, the practical comparisons, unified formulation, and synthetic tasks support the same conclusion: xLSTM's gains on tasks with complex dependencies stem from combining robust state tracking with counting-like accumulation.

Our contributions are: \textbf{(i)} \textbf{A comparison of leading subquadratic operators on tasks with complex dependencies.} We provide the first head-to-head comparison of xLSTM, Mamba-2, and Gated DeltaNet across settings that go beyond standard English-web pre-training. xLSTM backbones lead across most settings in empirical evaluation.
\textbf{(ii)} \textbf{A unified formulation of xLSTM, Mamba-2, and Gated DeltaNet that yields a hypothesis for the empirical differences.} We express the three backbones within a single framework, bridging their original state-space model and linear-attention notations. The formulation predicts that the architectures differ primarily on two primitives: accumulation and finite-state tracking.
\textbf{(iii)} \textbf{A validation of this hypothesis on synthetic tasks.} We empirically test the hypothesis on controlled length-generalization tasks for counting and state tracking.

\section{Experiments with Complex Dependencies}
We begin with the empirical comparison.
The goal of this section is not to explain why the architectures differ, but to establish the practical pattern that the rest of the paper explains.
We find that across code-focused language-model pre-training (Section \ref{sec:pretraining}), code-focused Transformer distillation (Section \ref{sec:distillation}), and time-series foundation-model pre-training (Section \ref{sec:timeseries}), xLSTM-family backbones outperform the other subquadratic operators in nearly all comparisons.
The strongest gains appear on complex structured tasks, while broad reasoning and commonsense benchmarks show the same direction with smaller margins.
This consistent empirical advantage motivates our architectural analysis in Section~\ref{sec:framework}, followed by controlled synthetic tasks in Section~\ref{sec:synthetic}.

\subsection{Code-focused Language-Model Pre-training}
\label{sec:pretraining}
We first compare subquadratic backbones in code-focused language-model pre-training.
Code is a complex language setting because it combines natural-language-like token distributions with formal syntax, variable binding, and executable structure.
We therefore use code generation as the primary metric in this subsection and report broad reasoning and commonsense tasks as a secondary check.

\paragraph{Experimental setup.}
We pretrain 400M-parameter inter-layer hybrid language models with \texttt{lm-engine}~\citep{mishra24lm_engine}, into which we integrate the xLSTM backbone alongside the existing Attention, Mamba-2, and Gated DeltaNet operators.
Here, ``inter-layer hybrid'' means that most layers use the tested subquadratic sequence operator, while a small number of layers remain standard self-attention layers.
We compare Gated DeltaNet, Mamba-2, and xLSTM\,$[7\!:\!1]$\footnote{Unlike the pure xLSTM[m:s] convention, the hybrid pre-training blocks also contain softmax attention. We fold these into the first index, so xLSTM[7:1] reads as 7 non-recurrent layers (6 mLSTM + 1 self-attention) to 1 recurrent layer.}.
All three models use 24 layers in total and keep three self-attention layers, matching the small fraction of self-attention used in contemporary hybrid architectures~\citep{lenz25jamba,ren25samba,nvidia25nemotron3nano,merrill26olmo_hybrid,team25kimi}.
We train under three data configurations: Nemotron-CC-Code-v1 \citep{nvidia25nemotron3nano} for 20B as well as 100B tokens, and a Nemotron-CC-Code-v1 + FineWeb-Edu mixture for 20B tokens.
We evaluate code generation with HumanEval pass@$k$ for $k\in\{2,8,16,64\}$, and report reasoning and commonsense accuracy on HellaSwag, PIQA, ARC-Easy, ARC-Challenge, and WinoGrande.
Further details on the setup are provided in Appendix \ref{app:lm-pretraining-implementation-details}.

\begin{figure}[t]
    \centering
    \includegraphics[width=0.8\linewidth]{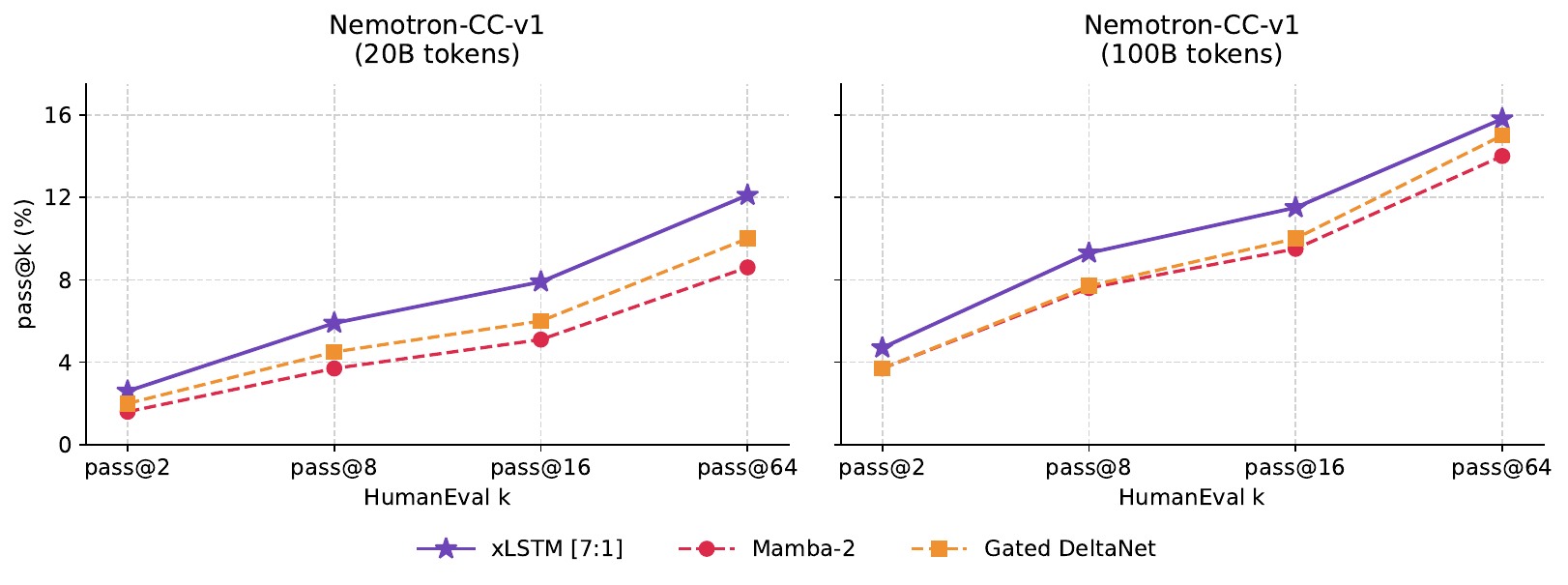}
    \caption{\textbf{HumanEval pass@$k$ after code-focused pre-training.}
    Results for 400M-parameter hybrid language models trained under the matched pre-training recipe on two data configurations:
    Nemotron-CC-Code-v1 for 20B tokens, Nemotron-CC-Code-v1 for 100B tokens. For 100B tokens, the gap between the different subquadratic backbones shrinks.}
    \label{fig:codeHumaneval}
\end{figure}

\paragraph{xLSTM\,$[7\!:\!1]$ consistently leads on code generation.}
As shown in Figure~\ref{fig:codeHumaneval}, xLSTM\,$[7\!:\!1]$ leads at every pass@$k$ and in every training configuration.
At pass@64, it improves over the next-best backbone by 1.43 points at 20B code tokens, 0.90 points at 100B code tokens, and 1.81 points on the mixed code-and-FineWeb-Edu corpus.
The runner-up is consistently Gated DeltaNet when trained on code-only data corpus.
Full results are reported in Appendix~\ref{app:pretraining} Tables~\ref{tab:humaneval_64_20b_code}, \ref{tab:humaneval_64_100b_code}, and~\ref{tab:humaneval_64_20b_fineweb_code}.

\paragraph{xLSTM\,$[7\!:\!1]$ keeps a smaller aggregate lead on reasoning and commonsense.}
Across the five reasoning and commonsense benchmarks, xLSTM\,$[7\!:\!1]$ has the best aggregate score in all three training configurations.
The margins are smaller than on HumanEval: it leads the closest non-xLSTM backbone by under 0.1 points at 20B and 100B code tokens, and by roughly half a point on the Nemotron-CC-Code-v1 + FineWeb-Edu mix.
Thus, the broad benchmarks agree with the code-generation ordering, but they make the xLSTM advantage less visible.
The full per-task results are reported in Appendix~\ref{app:pretraining} Tables~\ref{tab:benchmarks_pretraining_code_20B}, \ref{tab:benchmarks_pretraining_code_100B}, and~\ref{tab:benchmarks_pretraining}.

\paragraph{Discussion.}
The pre-training results show a consistent advantage for xLSTM\,$[7\!:\!1]$ among the compared backbones.
It leads to code generation in every training configuration and at every reported sampling budget.
It also has the best aggregate reasoning and commonsense score, although margins are smaller.
This supports the role of complex structured tasks in our evaluation: they reveal the same ordering as broad benchmarks, but with clearer separation between subquadratic backbones.
Section~\ref{sec:distillation} and Section~\ref{sec:timeseries} test whether xLSTM's advantage persists in distillation and time-series foundation-model pre-training.

\subsection{Code-focused Transformer Distillation into Subquadratic Students}
\label{sec:distillation}

Having compared subquadratic backbones in code-focused pre-training, we next ask whether these operators remain effective when initialized from a strong attention-based teacher.
Linearization, a form of knowledge distillation~\citep{hinton15distilling,mercat24linearizing}, converts an open-weight Transformer teacher into a subquadratic student and avoids a separate from-scratch pre-training run for each candidate operator.
We use the recipe of~\citet{hauzenberger26effective}, which combines a sliding-window-attention scaffold with attention sinks, hidden-state alignment, and sparse top-$k$ knowledge distillation.
Prior linearization work typically fixes the target operator family, such as linear/sliding-window attention, Mamba-style state-space mixers, RWKV, or gated recurrent structures~\citep{zhang25lolcats,bick24mohawk,bick25llamba,wang24mamba,goldstein25radlads,lan25liger}.
We instead compare xLSTM\,$[1\!:\!0]$ and Gated DeltaNet as plug-in matrix-state replacements under the same teacher, data, initialization scheme, and optimization recipe.
For code distillation, we additionally evaluate Gated DeltaNet\,$[-1,1]$, which uses the negative-eigenvalue parameterization of~\citet{grazzi25unlocking}.

\paragraph{Experimental setup.}
Our teacher is Qwen3-4B-Instruct~\citep{yang25qwen3}.
The student keeps the teacher's width, depth, and tokenizer.
We replace every multi-head-attention block with an intra-layer hybrid block: the tested linear-attention operator runs in parallel with sliding-window attention of window size 512 and four sink tokens~\citep{xiao24efficient,beltagy20longformer}, and a learned data-dependent gate fuses the two paths.
The linear-attention branch inherits the teacher's $\vec{q}$, $\vec{k}$, and $\vec{v}$ projection weights at initialization.
We follow the two-stage protocol of~\citet{hauzenberger26effective}.
Stage I aligns per-layer student outputs with the teacher's attention outputs under an MSE loss.
Stage II minimizes $0.9\,\mathrm{CE} + 0.1\,\mathrm{KL}$ with top-$k=256$ sparse teacher distribution.
Sequence length is 4{,}096, and we train for 10{,}000 stage-II optimization steps.
Code distillation uses Nemotron-Pretraining-Code-v2~\citep{nvidia25nemotron3nano}.
Because the distilled students keep the 4B teacher architecture, we use a harder code-generation suite than in Section~\ref{sec:pretraining}: HumanEval, HumanEval+, MBPP, and MBPP+.
Appendix~\ref{app:distillation-implementation-details} gives the full implementation details.

\paragraph{xLSTM\,$[1\!:\!0]$ and Gated DeltaNet are the plug-in comparison.}
The distillation setup supports linear-attention replacements that expose query, key, and value projections, admit chunkwise-parallel kernels, and reuse the teacher's attention projections at initialization.
Both xLSTM\,$[1\!:\!0]$ (mLSTM;~\citealp{beck24xlstm}) and Gated DeltaNet satisfy these constraints and slot in without changing the surrounding hybrid block.
Gated DeltaNet also provides the stricter non-xLSTM comparison for code distillation, since in the directly related code-only pre-training setting, it is the strongest non-xLSTM backbone on HumanEval at both 20B and 100B tokens (Section~\ref{sec:pretraining}; Appendix Tables~\ref{tab:humaneval_64_20b_code} and~\ref{tab:humaneval_64_100b_code}).
In contrast, xLSTM\,$[0\!:\!1]$ is sequential and has no query-key-value analogue to initialize from the teacher, while Mamba-2 ties its input and forget gates through parameters that do not map directly to teacher attention weights.
We therefore use this experiment as a controlled comparison between plug-in matrix-state operators, not as a test of the full xLSTM hybrid family.\footnote{The distillation setup can extend to xLSTM\,$[0\!:\!1]$ or other xLSTM\,$[m\!:\!s]$ variants, and Mamba-2, but these require additional initialization and architecture choices and would no longer isolate the plug-in matrix-state comparison studied here.}
We report Gated DeltaNet with its default parameterization and, for code, the Gated DeltaNet\,$[-1,1]$ variant (see Appendix~\ref{app:distillation-implementation-details}).

\paragraph{xLSTM\,$[1\!:\!0]$ gives the stronger code student on average.}
Across the four code benchmarks at pass@1, xLSTM\,$[1\!:\!0]$ matches or exceeds default Gated DeltaNet on three metrics and trails only on MBPP+ by 0.014 (Table~\ref{tab:distill-code}).
The negative-eigenvalue variant improves Gated DeltaNet on HumanEval and HumanEval+, but trails the default variant on MBPP and MBPP+.
The average across HumanEval, HumanEval+, MBPP, and MBPP+ is 0.755 for default Gated DeltaNet, 0.756 for Gated DeltaNet\,$[-1,1]$, and 0.768 for xLSTM\,$[1\!:\!0]$.
The full HumanEval and HumanEval+ pass@$k$ spread in Appendix~\ref{app:distill-code-passk} shows that xLSTM\,$[1\!:\!0]$ remains strongest across sampling budgets.

\paragraph{Discussion.}
The distillation experiment complements code-focused pre-training by testing the same operator comparison inside a Transformer-linearization pipeline.
Under a fixed teacher, data, initialization scheme, and optimization recipe, xLSTM\,$[1\!:\!0]$ gives the stronger code student on average.
This shows that the xLSTM advantage in code-focused settings does not rely only on the recurrent layers used in xLSTM inter-layer hybrids (Section \ref{sec:pretraining}): the linear-attention component is already a strong plug-in operator.
Appendix~\ref{app:distill-math} reports the corresponding math-distillation results, where xLSTM\,$[1\!:\!0]$ also leads the aggregate while Gated DeltaNet remains slightly stronger on MATH-500.
Together, the code and math results support xLSTM\,$[1\!:\!0]$ as an effective matrix-state replacement in Transformer distillation to subquadratic architectures.

\begin{table}[t]
\centering
\small
\setlength{\tabcolsep}{6pt}
\caption{\textbf{Code distillation results at pass@1.}
Students are distilled from Qwen3-4B-Instruct.
xLSTM\,$[1\!:\!0]$ leads on three of four benchmarks and on average, while the default Gated DeltaNet performs better on MBPP+.
Gated DeltaNet\,$[-1,1]$ improves over the default Gated DeltaNet on HumanEval and HumanEval+ but not on MBPP and MBPP+.
Higher is better; the best student result per column is shown in \textbf{bold}.}
\begin{tabular}{lccccc}
\toprule
Student & HumanEval $\uparrow$ & HumanEval+ $\uparrow$ & MBPP $\uparrow$ & MBPP+ $\uparrow$ & Avg.\ $\uparrow$ \\
\midrule
Qwen3-4B-Instruct (teacher) & 0.914 & 0.835 & 0.708 & 0.847 & 0.826 \\
\midrule
xLSTM\,$[1\!:\!0]$          & \textbf{0.831} & \textbf{0.764} & \textbf{0.689} & 0.788 & \textbf{0.768} \\
Gated DeltaNet              & 0.802 & 0.739 & 0.677 & \textbf{0.802} & 0.755 \\
Gated DeltaNet\,$[-1,1]$    & 0.813 & 0.745 & 0.671 & 0.796 & 0.756 \\
\bottomrule
\end{tabular}
\label{tab:distill-code}
\end{table}

\subsection{Time-series Foundation-Model Pre-training}
\label{sec:timeseries}

After code-focused pre-training and distillation, we ask whether the same architectural comparison also holds outside these tasks.
\ac{tsfm} have so far been built primarily on Transformer backbones~\citep{ansari24chronos,woo24unified,das24timesfm,cohen24toto}, with subquadratic backbones emerging as recent alternatives.
TiRex~\citep{auer25tirex} demonstrates that an xLSTM-based \ac{tsfm} is competitive, TempoPFN~\citep{moroshan25tempopfn} adopts Gated DeltaProduct \citep{siems25deltaproduct}, a Gated DeltaNet variant, and FlowState~\citep{graf25flowstate} uses the S5 state-space variant.
These works establish strong individual designs, but they do not compare subquadratic backbone families under a matched setting.
Time series, therefore, provides a complementary complex-task setting with continuous values, heterogeneous domains and frequencies, and forecasting horizons that require models to use information from long histories.

\paragraph{Experimental setup.}
We use the time-series pre-training protocol of~\citet{auer25tirex}: the same corpus, patching scheme, optimizer, and forecasting head are shared across all models, while only the sequence mixer changes.
Thus, the comparison isolates the backbone choice within a fixed forecasting pipeline.
We compare Mamba-2, Gated DeltaNet, and xLSTM\,$[3\!:\!1]$.
Models are trained at five parameter scales: 1M, 4M, 10M, 40M, and 80M parameters, with width and depth chosen to match the parameter count in each setting.
This range is small compared with contemporary language models, but it is a practical scale range for \acp{tsfm}; for example, both TiRex and FlowState report strong forecasting performance in the 20-35M parameter range~\citep{auer25tirex, graf25flowstate}.
We evaluate zero-shot on GIFT-Eval~\citep{aksu24gift-eval}, a heterogeneous forecasting benchmark spanning multiple domains and frequencies, and report \ac{mase} and \ac{crps} aggregated by geometric mean.
Appendix~\ref{app:tsfm-implementation-details} gives the implementation details, and Appendix~\ref{app:timeseries-results} reports the full numerical results.

\begin{figure}[h]
    \centering
    \includegraphics[width=0.75\linewidth]{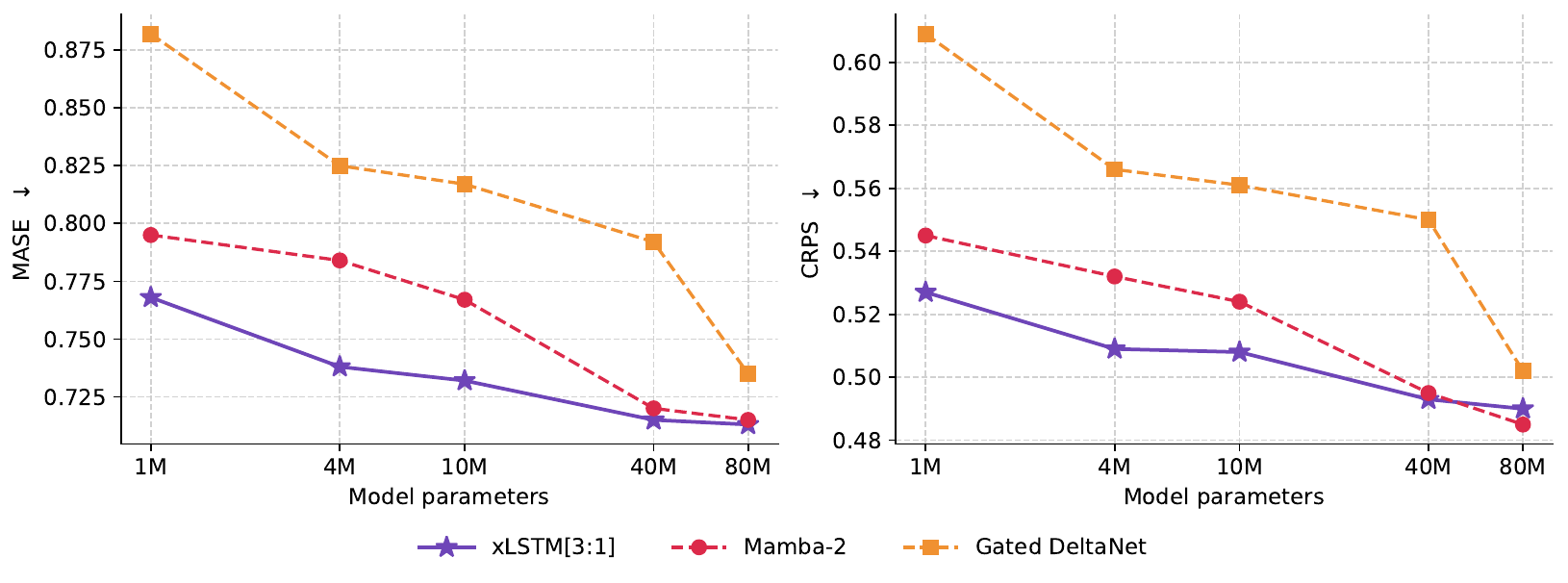}
    \caption{\textbf{GIFT-Eval performance of \ac{tsfm} over five parameter scales.} MASE and CRPS scores (lower is better) for matched training recipe. xLSTM architectures provide the best scores, with the gap narrowing as the parameter scale grows.}
    \label{fig:tsfm-results}
\end{figure}

\paragraph{xLSTM\,$[3\!:\!1]$ leads from 1M to 40M parameters.}
Figure \ref{fig:tsfm-results} sweep shows a clear small-to-mid-scale advantage for xLSTM\,$[3\!:\!1]$.
It achieves the best \ac{mase} and \ac{crps} at every scale from 1M to 40M parameters.
The separation is most visible at small and mid scales.
At 10M parameters, for example, xLSTM\,$[3\!:\!1]$ reaches 0.733 \ac{mase} and 0.508 \ac{crps}, compared with 0.767 and 0.525 for the next-best model (Mamba-2).
At 80M parameters, the models nearly converge: xLSTM\,$[3\!:\!1]$ and Mamba-2 are on par on \ac{mase}, while Mamba-2 slightly leads on \ac{crps} by 0.005.
The full values are reported in Appendix~\ref{app:timeseries-results}, Table~\ref{tab:timeseries-results}.

\paragraph{Discussion.}
The time-series results extend the practical comparison beyond language.
Under a matched \ac{tsfm} recipe, xLSTM\,$[3\!:\!1]$ is the strongest backbone from 1M to 40M parameters, while the gap narrows at 80M.
Together with Sections~\ref{sec:pretraining} and~\ref{sec:distillation}, these results show a consistent empirical pattern: xLSTM-family backbones outperform the competing subquadratic operators in nearly all matched practical comparisons.
The few exceptions are narrow: Gated DeltaNet leads MBPP+ in distillation, and Mamba-2 leads \ac{crps} by 0.005 at 80M parameters.
The next sections hypothesize where this advantage comes from.
Section~\ref{sec:framework} analyzes the memory dynamics of the architectures, and Section~\ref{sec:synthetic} tests the resulting hypotheses on controlled counting and state-tracking tasks.

\section{Analysis of Leading Subquadratic Attention Architectures}
\label{sec:framework}

In order to better understand the empirical differences between the architectures, we take a closer look at the underlying attention mechanisms.
Concretely, we express xLSTM, Mamba-2, and Gated DeltaNet in terms of linear attention with a gating mechanism using a common notation.

Attention mechanisms \citep{vaswani17attention, bahdanau2015neural} are specified in terms of \emph{query}, \emph{key}, and \emph{value} matrices,
\begin{align*}
    \mat{Q} &= \mat{X} \mat{W}_\mathrm{q}^\T &
    \mat{K} &= \mat{X} \mat{W}_\mathrm{k}^\T &
    \mat{V} &= \mat{X} \mat{W}_\mathrm{v}^\T,
\end{align*}
where $\mat{W}_{\{\mathrm{q}, \mathrm{k}\}} \in~\reals^{D_\mathrm{qk} \times D}, \mat{W}_\mathrm{v} \in~\reals^{D_\mathrm{v} \times D}$ represent learnable parameters of the affine projections, and $\mat{X} \in~\reals^{T \times D}$ is a sequence of inputs.
With these matrices, the regular, causal softmax attention can be written as
$\softmax(\mat{Q} \mat{K}^\T \odot \mat{M}) \mat{V},$
where $\mat{M}~\in~\{-\infty,1\}^{T \times T}$ is a causal masking matrix, such that $m_{ij} = 1$ if $i \geq j$ and $-\infty$ otherwise.

\paragraph{Linear attention}\citep{katharopoulos20transformers} is a basic subquadratic attention variant that does away with the softmax in regular attention.
This enables a recurrent formulation for single-head linear attention that exposes an explicit matrix state, $\mat{C} \in \reals^{D_\mathrm{qk} \times D_\mathrm{v}}$:
\begin{align*}
    \mat{H} &= \frac{(\mat{Q} \mat{K}^\T \odot \mat{M})\mat{V}}{\big(|\mat{Q} \mat{K}^\T| \odot \mat{M}\big) \mat{1}} \tag{parallel} \\
    \vec{h}_t &= \frac{\vec{q}_t \mat{C}_{t}}{|\vec{q}_t \vec{n}_t|} &
    \mat{C}_t &= \mat{C}_{t-1} + \vec{k}_t \otimes \vec{v}_t &
    \vec{n}_t &= \vec{n}_{t-1} + \vec{k}_t, \tag{recurrent}
\end{align*}
where division is element-wise, $\otimes$ denotes the outer product, $1 \leq t \leq T$, $\mat{C}_0 = \mat{0}$, $\vec{n}_0 = \vec{0}$ and $\mat{M} \in \{0,1\}^{T \times T}$ is a causal masking matrix.
Note that the explicit normalization, inspired by the normalization inside the softmax function \citep{katharopoulos20transformers}, is commonly ignored in practice \citep[cf.][]{yang24gla, yang26fla}.
Instead, this normalization is implemented by normalization layers \citep[e.g.][]{ba16layernorm, wu18groupnorm}.
To ease notation, we omit the explicit normalization throughout this work, unless necessary.

The recurrent formulation of linear attention is what enables its subquadratic nature, i.e., $\mathcal{O}(T)$ instead of $\mathcal{O}(T^2)$ for regular attention and the parallel formulation.
However, the recurrent formulation can not make use of hardware that is optimized for matrix multiplications, making it slow in practice.
As a result, practical implementations use the intermediate chunk-wise formulation \citep{hua22transformer}:
\begin{equation*}
    \begin{aligned}
        \mat{H}_{[n]} &= (\mat{Q}_{[n]} \mat{K}_{[n]}^\T \odot \mat{M}) \mat{V}_{[n]} + \mat{Q}_{[n]} \mat{C}_{(n-1)C} \\
        \mat{C}_{nC} &= \mat{C}_{(n-1)C} + \mat{K}_{[n]}^\T \mat{V}_{[n]},
    \end{aligned}
    \tag{chunkwise}
\end{equation*}
where the $C$ time steps within each chunk are processed in parallel, while the explicit state is used to connect the different chunks sequentially.
This enables linear complexity while allowing efficient usage of modern hardware.
Here, $1 \leq n \leq \lceil T / C \rceil$, $\mat{X}_{[n]}$ is a short-hand for $\mat{X}_{((n-1)C+1:nC)}$, i.e.~the chunk for all time-steps $t$ with $(n-1)C + 1 \leq t \leq nC$ \citep[cf.~][]{yang24gla}, and, with slight abuse of notation, $\mat{M} \in \{0,1\}^{C \times C}$ represents the causal mask for a single chunk.
Note that the chunk-wise formulation reduces to the (unnormalized) parallel and recurrent formulations if $C = T$ and $C = 1$, respectively.

\paragraph{xLSTM} \citep{beck24xlstm} is a modern version of {LSTM} \citep{hochreiter97lstm, gers99learning}.
It consists of a linear attention component, called {mLSTM} or {xLSTM}[1:0], and a non-linear recurrent component, called {sLSTM} or {xLSTM}[0:1].
These components can be combined to form the {xLSTM}[$m$:$s$] architecture, where $m$ and $s$ represent the number of linear attention and recurrent layers, respectively.
The recurrence of a single {xLSTM}[0:1] head is given by:
\begin{equation}
    \begin{aligned}
        \vec{v}_t &= {\color{xlstm_accent} \tanh}(\mat{W}_\mathrm{v} \vec{x}_t + {\color{xlstm_accent} \mat{R}_\mathrm{v} \vec{h}_{t-1}}) &
        \vec{q}_t &= \vec{e}_1 \qquad \vec{k}_t = \vec{1} \\
        \vec{i}_t &= {\color{xlstm_accent} \exp}(\mat{W}_\mathrm{i} \vec{x}_t + {\color{xlstm_accent} \mat{R}_\mathrm{i} \vec{h}_{t-1}}) &
        \vec{f}_t &= \sigma(\mat{W}_\mathrm{f} \vec{x}_t + {\color{xlstm_accent} \mat{R}_\mathrm{f} \vec{h}_{t-1}}) \\
        \vec{c}_t &= {\color{xlstm_accent} \diag}(\vec{f}_t) \, \vec{c}_{t-1} + {\color{xlstm_accent} \diag}(\vec{i}_t) \, \vec{v}_t &
        \vec{n}_t &= \diag(\vec{f}_t) \, \vec{n}_{t-1} + \vec{i}_t \\
        \vec{h}_t &= \frac{\vec{c}_t}{\vec{n}_t},
    \end{aligned}
    \label{eq:sLSTM}
\end{equation}
where the division is element-wise, and $\mat{W}_{\{\mathrm{i},\mathrm{f},\mathrm{v}\}}$ and $\mat{R}_{\{\mathrm{i},\mathrm{f},\mathrm{v}\}} \in \reals^{D \times D}$ are learnable parameters for the input and forget gate, as well as the state update.
Note that we redefine the keys, queries, and values and explicitly model the normalizer state because the xLSTM[0:1] does not perfectly align with the linear attention paradigm.
The linear attention mechanism of a single {xLSTM}[1:0] head, on the other hand, can be expressed using the following recurrence:
\begin{equation}
    \begin{aligned}
        i_t &= {\color{xlstm_accent} \exp}(\vec{w}_\mathrm{i} \vec{x}_t) &
        f_t &= \sigma(\vec{w}_\mathrm{f} \vec{x}_t) \\
        \vec{h}_t &= \vec{q}_t \mat{C}_t &
        \mat{C}_t &= f_t \, \mat{C}_{t-1} + i_t \, \vec{k}_t \otimes \vec{v}_t.
    \end{aligned}
    \label{eq:mLSTM}
\end{equation}
Here, $\vec{w}_{\{\mathrm{i},\mathrm{f}\}}$ are the learnable parameters for the input and forget, and we assume biases are implicit.
One of the key differences between xLSTM and LSTM is the exponential input gate.
When normalized correctly, this input gate behaves like a softmax over time, allowing the model to down-weight, or overwrite, previous values when the current value is more important.
The main difference between xLSTM[0:1] and xLSTM[1:0] is the use of recurrent weights to incorporate the previous state.
This recurrence enables state-tracking capabilities similar to those of other recurrent networks \citep{merrill19sequential}.
Note that we ignore the output gate in these formulations, as it is typically implemented using a {SwiGLU} \citep{shazeer20glu}, which has become a common component in the block-wrappers around the core attention mechanism \citep{gu24mamba,yang24gdn,beck24xlstm}.

\paragraph{Mamba-2} \citep{dao24mamba2} is a linear attention variant derived from state-space models \citep[e.g.][]{gu22s4,gupta22diagonal,gu24mamba}.
The recurrent formulation for a single head in the attention mechanism of Mamba-2 can be expressed as follows:
\begin{equation}
    \begin{aligned}
        i_t &= {\color{mamba2_accent} \softplus}(\vec{w}_{\color{mamba2_accent} \Delta} \vec{x}_t) &
        f_t &= \big(1 - \sigma(\vec{w}_{\color{mamba2_accent} \Delta} \vec{x}_t)\big)^{\color{mamba2_accent} a} \\
        \vec{h}_t &= \vec{q} \mat{C}_t &
        \mat{C}_t &= f_t \, \mat{C}_{t-1} + i_t \, \vec{k}_t \otimes \vec{v}_t.
    \end{aligned}
    \label{eq:mamba2}
\end{equation}
Here, $\vec{w}_\Delta$ are the learnable parameters for computing the sample time in the zero-order hold discretisation \citep{gupta22diagonal,gu24mamba}, and $a \in \reals_{\geq 0}$ is a non-negative learned parameter to construct the 1-SS transition matrix.
As a result, Mamba-2 can be interpreted as an {xLSTM}[1:0] with tied input and forget gates, making it similar to a \ac{gru} \citep{cho14gru,dao24mamba2}.
Because \acp{gru} are known to have issues with counting \citep[e.g.][]{weiss18practical}, we expect Mamba-2 to have similar limitations.

\paragraph{Gated DeltaNet} \citep{yang24gdn} is practically a combination of the fast-weight mechanism of Delta-Nets \citep{schlag21deltanet} and the 1-SS transition dynamics of Mamba-2 \citep{dao24mamba2}.
The recurrence of this linear attention variant can be written as:
\begin{equation}
    \begin{aligned}
        i_t &= \sigma(\vec{w}_\beta \vec{x}_t) &
        f_t &= \big(1 - \sigma(\vec{w}_\alpha \vec{x}_t)\big)^{\color{gdn_accent} a} \\
        \vec{h}_t &= \frac{\vec{q}_t}{\norm{\vec{q}_t}} \mat{C}_t &
        \mat{C}_t &= f_t {\color{gdn_accent} \Big(\mat{I} - i_t \frac{\vec{k}_t \otimes \vec{k}_t}{\norm{\vec{k}_t}^2}\Big)} \mat{C}_{t-1} + i_t \, \frac{\vec{k}_t}{\norm{\vec{k}_t}} \otimes \vec{v}_t,
    \end{aligned}
    \label{eq:gdn}
\end{equation}
where $\vec{w}_\alpha$ and $\vec{w}_\beta$ are the learnable parameters for the gating and write-strength \citep{schlag21deltanet}, respectively, and $a \in \reals_{\geq 0}$ is a non-negative learned parameter from Mamba-2.
We note that Gated DeltaNet can be interpreted as an xLSTM[1:0] with an additional state transformation.
The matrix $\mat{I} - \frac{\vec{k}_t \otimes \vec{k}_t}{\norm{\vec{k}_t}^2}$ is an orthogonal projection onto the null-space of $\vec{k}_t$.
This means that the additional transformation removes all components in the direction of $\vec{k}_t$ from the state when $i_t = 1$.
E.g., when $\vec{k}_t = \vec{k}_s$ for some $s < t$, the old value, $\vec{v}_s$, will be removed from the state matrix and replaced by the new value, $\vec{v}_t$.
Because old values are always overwritten, Gated DeltaNets are also expected to have problems with counting.

\paragraph{All linear attention variants exhibit very similar gating mechanisms.}
Concretely, each of these models can be written in terms of input and forget gates, similar to LSTM.
Whereas xLSTM and Gated DeltaNets have independent gates, the input and forget gate in Mamba-2 are tied and therefore expected to be less expressive.
Another key difference lies in the overwriting mechanism of the input gate.
Mamba-2 has limited capabilities to correct weights in previous time-steps due to its linear-like input gate.
Gated DeltaNet explicitly overwrites old values in the state, making it better suited for retrieval tasks \citep{yang24gdn}, but can be problematic for counting.
The xLSTM architecture enables the most flexible weighting correction that allows down-weighting old values by means of the softmax-like input gate.

\paragraph{We attribute this advantage to xLSTM's ability to solve counting and state tracking.}
Recent theory points to two capabilities that sequence models often fail to combine: accumulation over unbounded lengths and finite-state tracking~\citep{weiss18practical, merrill26why}.
Mamba-style state-space models and (Gated) DeltaNets inherit the $\mathrm{TC}^0$ ceiling of Transformers and cannot solve hard state-tracking problems such as permutation composition~\citep{merrill24illusion, grazzi25unlocking}; related limitations appear in code modeling~\citep{siems26learning}.
However, \citet{grazzi25unlocking} points out that this can be alleviated by enabling negative eigenvalues in the state-transition matrix.
Concretely, mapping the forget gate to values in $[-1, 1]$ instead of $[0,1]$ should improve state-tracking significantly.
Within xLSTM, the matrix-state update provides a natural mechanism for accumulation, while the nonlinear recurrent component can support structured state updates.
This makes mixed xLSTM\,$[m\!:\!s]$ architectures a plausible way to combine accumulation with state tracking in a scalable backbone.
We therefore interpret the strong performance of xLSTM on complex domains such as code and time series as evidence that these capabilities are useful in combination, rather than as a consequence of either mechanism in isolation.

\section{Experiments on Accumulation and State Tracking}
\label{sec:synthetic}
\begin{figure}[h]
    \centering
    \includegraphics[width=0.8\linewidth]{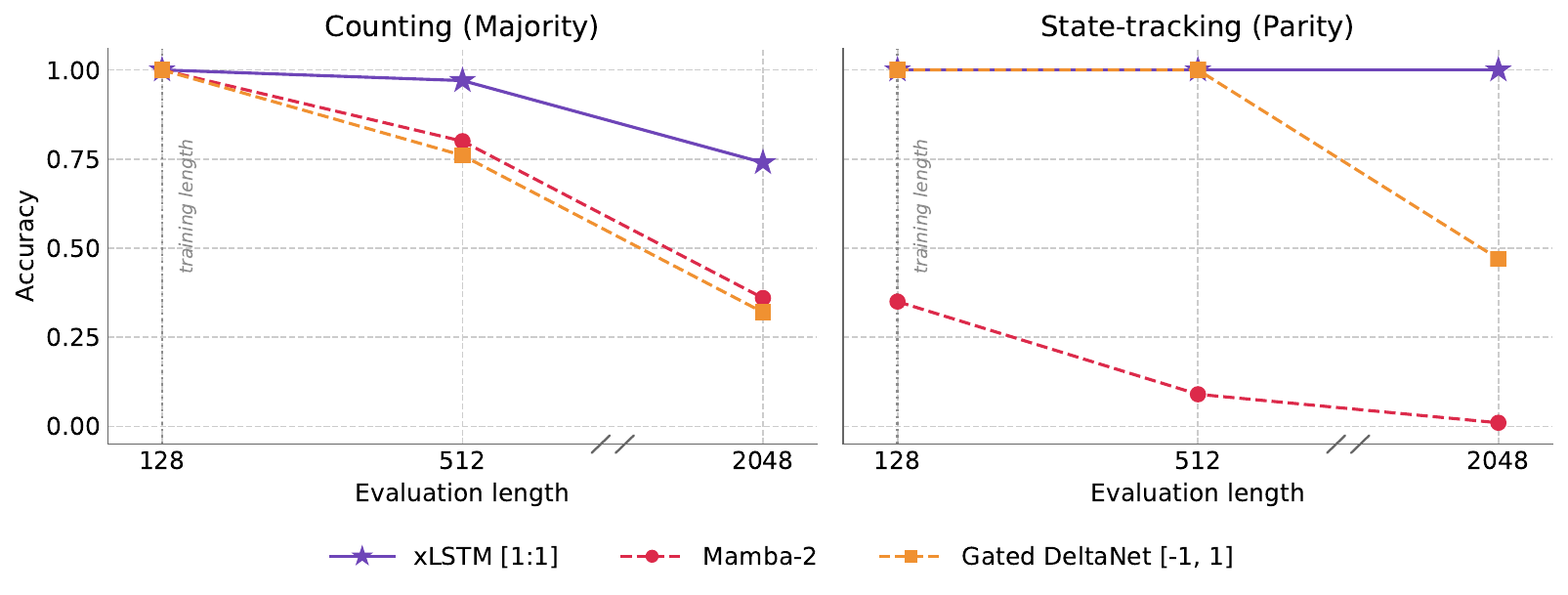}
    \caption{\textbf{Length generalization on accumulation and state-tracking.}
    Two representative tasks (Majority counting on the left, parity on the right) on which contemporary subquadratic designs diverge.
    Models are trained at length 128 (dotted line) and evaluated at 128, 512, and 2048; the break on the x-axis marks the $4\times$ jump from 512 to 2048.
    xLSTM$[1\!:\!1]$ is the only configuration that length-generalizes on both tasks: it achieves the highest counting accuracy at every length and solves parity perfectly throughout.
    Gated DeltaNet with the negative-eigenvalue parameterization of~\citet{grazzi25unlocking} solves parity in-distribution but drops to $0.47$ at length 2048; Mamba-2 never solves either.
    }
    \label{fig:synthetic}
\end{figure}

\paragraph{Experimental Setup.}
Each model is trained at sequence length 128 and evaluated at lengths 128, 512, and 2048, covering a $4\times$ and $16\times$ extrapolation step.
The counting tasks are $A^n B^n$, $A^n B^n C^n$, and Majority; the state-tracking tasks are Parity, Modular Arithmetic ($\mathbb{Z}_5$), and word-problem evaluation in the symmetric group $S_3$.
We compare Mamba-2, Gated DeltaNet with the default non-negative eigenvalue parameterization (Gated DeltaNet) and with the negative-eigenvalue parameterization of~\citet{grazzi25unlocking} (Gated DeltaNet\,$[-1,1]$), xLSTM[1:0] and xLSTM[1:1].
All models are trained under identical setups; full details are in Appendix~\ref{app:synthetic-implementation-details}.

\paragraph{Only xLSTM combines both accumulation and state tracking.}
Table~\ref{tab:synthetic-results} in Appendix \ref{app:synthetic-results} reports accuracy at each evaluation length, and Figure \ref{fig:synthetic} matches the architectural predictions above.
Mamba-2 collapses on every task: $A^n B^n$ accuracy drops from $1.000$ at length 128 to $0.241$ at 2048, and Parity accuracy never exceeds $0.352$ even in-distribution.
Default Gated DeltaNet solves the easiest counting variant at moderate length, but degrades to $0.268$ on Majority at 2048 and never solves any state-tracking task.
Gated DeltaNet\,$[-1,1]$ recovers Parity and $S_3$ in-distribution ($1.000$ at length 128). Still, length-generalisation on these tasks is only partial: Parity drops to $0.472$ at length 2048 and $ S_3$ to $0.667$, and the parameterization does not improve modular arithmetic ($0.452$ at length 2048).
xLSTM[1:0] length-generalises on every counting task ($0.892$ on $A^n B^n$, $0.932$ on $A^n B^n C^n$, $0.763$ on Majority at length 2048) but, as expected for a linear-attention block, fails on every state-tracking task.
The hybrid xLSTM$[1\!:\!1]$ is the best balanced configuration rather than the best model on every task. It is exact on all three state-tracking tasks and retains useful counting extrapolation, but it trails xLSTM$[1\!:\!0]$ on the hardest long counting tasks. 

\paragraph{Discussion.}
The synthetic results separate the two primitives.
The xLSTM\,$[1\!:\!0]$ is the strongest counting model, consistent with the accumulation mechanism introduced above, but fails on state-tracking tasks.
Conversely, adding the recurrent xLSTM\,$[0\!:\!1]$ component in xLSTM\,$[1\!:\!1]$ yields perfect state tracking at all tested lengths, while preserving useful but weaker counting extrapolation.
Gated DeltaNet exhibits the expected tradeoff: the default parameterization does not solve state tracking, while the negative-eigenvalue variant improves state tracking but does not consistently extrapolate.
These results motivate m-heavy xLSTM\,$[m\!:\!s]$ models in practical settings, such as xLSTM$[7\!:\!1]$ for language pretraining and xLSTM$[3\!:\!1]$ for time series: they retain the efficient accumulation block as the dominant component while adding a smaller number of recurrent layers for state tracking.

\section{Conclusion}

We have conducted the first comparison of xLSTM, Mamba-2, and Gated DeltaNet across more complex data domains.
Our experiments show that xLSTM backbones outperform other subquadratic operators in nearly all comparisons across all studied settings.
These empirical results motivated our analysis of why xLSTM performs better under the proposed settings.
To explain xLSTM's advantage on complex tasks, we derived a common formulation that makes the architectures directly comparable at the level of memory dynamics.
Our formulation shows how gates, normalization, and overwriting mechanisms shape two primitive capabilities: \textit{accumulation} and \textit{finite-state tracking}.
We have found that Mamba-2 couples writing and forgetting through tied gates, while Gated DeltaNet adds explicit overwriting, which helps replacement-style memory updates but can interfere with accumulation.
In contrast, xLSTM separates matrix-state linear attention from recurrent state updates, giving it a direct way to combine accumulation, state tracking, and flexible memory correction.

\paragraph{Limitations and Future Work.}
First, our code-focused language modeling is conducted at a 400M-parameter scale, and our distillation pipeline uses a single teacher; only the time-series experiments include a scaling sweep.
Extending the comparison to larger model scales, additional teachers, and further data domains is a natural next step.
Second, we focus on recent leading subquadratic architectures and exclude families already compared in \citet{beck24xlstm}; a broader operator survey under our unified formulation would further sharpen the picture.

\section*{Acknowledgments}
This work was supported by the Austrian Science Fund (FWF) 10.55776/COE12 and the European Union’s Horizon Europe research and innovation program under grant agreement number 101214398 (ELLIOT).
The ELLIS Unit Linz, the LIT AI Lab, and the Institute for Machine Learning are supported by the Federal State of Upper Austria.
We acknowledge the EuroHPC Joint Undertaking for awarding us access to Leonardo at CINECA, Italy.
\small
\bibliographystyle{plainnat}
\bibliography{refs}
\normalsize

\clearpage
\appendix
\startcontents[appendix]

\section*{Appendix}
\printcontents[appendix]{l}{1}{\setcounter{tocdepth}{2}}
\clearpage

\clearpage

\section{Notation table}

\begin{longtable}{@{}lp{7.5cm}l@{}}
\caption{Notation used in this paper.}
\label{tab:notation} \\

\toprule
\textbf{Symbol} & \textbf{Definition} & \textbf{Type} \\
\midrule
\endfirsthead

\multicolumn{3}{c}{\tablename\ \thetable{} -- continued} \\
\toprule
\textbf{Symbol} & \textbf{Definition} & \textbf{Type} \\
\midrule
\endhead

\bottomrule
\endlastfoot

\multicolumn{3}{@{}l}{\textit{General sequence notation}} \\
\midrule
$t$                       & Time step index                                              & scalar \\
$T$                       & Total sequence length                                        & scalar \\
$n$                       & Chunk index, $1 \leq n \leq \lceil T/C \rceil$               & scalar \\
$C$                       & Chunk size                                                   & scalar \\
$D$                       & Input / model dimensionality                                 & scalar \\
$D_\mathrm{qk}$           & Query and key dimensionality                                 & scalar \\
$D_\mathrm{v}$            & Value dimensionality                                         & scalar \\
$\vec{x}_t$               & Input vector at time step $t$                                & $\reals^{D}$ \\
$\mat{X}$                 & Input sequence matrix                                        & $\reals^{T \times D}$ \\
$\mat{H}$                 & Output sequence matrix                                       & $\reals^{T \times D_\mathrm{v}}$ \\

\addlinespace[0.5em]
\multicolumn{3}{@{}l}{\textit{Attention mechanism}} \\
\midrule
$\vec{q}_t$               & Query vector at time step $t$                                & $\reals^{D_\mathrm{qk}}$ \\
$\vec{k}_t$               & Key vector at time step $t$                                  & $\reals^{D_\mathrm{qk}}$ \\
$\vec{v}_t$               & Value vector at time step $t$                                & $\reals^{D_\mathrm{v}}$ \\
$\mat{Q}$                 & Stacked query matrix                                         & $\reals^{T \times D_\mathrm{qk}}$ \\
$\mat{K}$                 & Stacked key matrix                                           & $\reals^{T \times D_\mathrm{qk}}$ \\
$\mat{V}$                 & Stacked value matrix                                         & $\reals^{T \times D_\mathrm{v}}$ \\
$\mat{W}_\mathrm{q}$      & Query projection weight matrix                               & $\reals^{D_\mathrm{qk} \times D}$ \\
$\mat{W}_\mathrm{k}$      & Key projection weight matrix                                 & $\reals^{D_\mathrm{qk} \times D}$ \\
$\mat{W}_\mathrm{v}$      & Value projection weight matrix                               & $\reals^{D_\mathrm{v} \times D}$ \\
$\mat{M}$                 & Causal masking matrix, $m_{ij} = 1$ iff $i \geq j$           & $\{0,1\}^{T \times T}$ \\
$\mat{C}_t$               & Matrix cell state / linear-attention memory at time $t$      & $\reals^{D_\mathrm{qk} \times D_\mathrm{v}}$ \\
$\vec{n}_t$               & Normalizer state at time step $t$                            & $\reals^{D_\mathrm{qk}}$ \\

\addlinespace[0.5em]
\multicolumn{3}{@{}l}{\textit{xLSTM$[0\!:\!1]$ -- nonlinear recurrent component (sLSTM)}} \\
\midrule
$\vec{h}_t$               & Hidden state at time step $t$                                & $\reals^{D}$ \\
$\vec{c}_t$               & Cell state at time step $t$                                  & $\reals^{D}$ \\
$\vec{i}_t$               & Input gate vector at time step $t$ (exponential activation)  & $\reals^{D}$ \\
$\vec{f}_t$               & Forget gate vector at time step $t$ ($\sigma$ activation)    & $\reals^{D}$ \\
$\mat{W}_\mathrm{i}, \mat{W}_\mathrm{f}, \mat{W}_\mathrm{v}$
                          & Input weight matrices for input gate, forget gate, cell input & $\reals^{D \times D}$ \\
$\mat{R}_\mathrm{i}, \mat{R}_\mathrm{f}, \mat{R}_\mathrm{v}$
                          & Recurrent weight matrices for input gate, forget gate, cell input & $\reals^{D \times D}$ \\

\addlinespace[0.5em]
\multicolumn{3}{@{}l}{\textit{xLSTM$[1\!:\!0]$ -- linear attention component (mLSTM)}} \\
\midrule
$i_t$                     & Scalar input gate at time step $t$ (exponential activation)  & scalar $\in \reals$ \\
$f_t$                     & Scalar forget gate at time step $t$ ($\sigma$ activation)    & scalar $\in \reals$ \\
$\vec{w}_\mathrm{i}$      & Weight vector for input gate                                 & $\reals^{D}$ \\
$\vec{w}_\mathrm{f}$      & Weight vector for forget gate                                & $\reals^{D}$ \\

\addlinespace[0.5em]
\multicolumn{3}{@{}l}{\textit{xLSTM$[m\!:\!s]$ -- full architecture}} \\
\midrule
$m$                       & Number of linear-attention layers (xLSTM$[1\!:\!0]$ blocks)  & scalar \\
$s$                       & Number of nonlinear recurrent layers (xLSTM$[0\!:\!1]$ blocks) & scalar \\

\addlinespace[0.5em]
\multicolumn{3}{@{}l}{\textit{Mamba-2}} \\
\midrule
$\vec{w}_\Delta$          & Weight vector for sample-time (discretization) projection    & $\reals^{D}$ \\
$a$                       & Non-negative learned parameter for 1-SS transition matrix    & scalar $\in \reals_{\geq 0}$ \\

\addlinespace[0.5em]
\multicolumn{3}{@{}l}{\textit{Gated DeltaNet}} \\
\midrule
$\vec{w}_\alpha$ & Weight vector for forget gate               & $\reals^{D}$, scalar \\
$\vec{w}_\beta$   & Weight vector for write-strength gate       & $\reals^{D}$, scalar \\

\addlinespace[0.5em]
\multicolumn{3}{@{}l}{\textit{Activation functions and operators}} \\
\midrule
$\tanh$                   & Hyperbolic tangent; cell input activation in xLSTM$[0\!:\!1]$ & function \\
$\sigma$                  & Sigmoid function; forget and output gate activation          & function \\
$\exp$                    & Exponential function; input gate in xLSTM                   & function \\
$\softplus$               & Softplus function; input gate in Mamba-2                     & function \\
$\softmax$                & Softmax function; attention normalization                    & function \\
$\odot$                   & Element-wise (Hadamard) product                              & operator \\
$\otimes$                 & Outer product                                                & operator \\
$\norm{\vec{v}}$          & Euclidean norm of vector $\vec{v} \in \reals^{D}$            & scalar \\
$\diag(\vec{v})$          & Diagonal matrix with entries $\vec{v} \in \reals^{D}$        & $\reals^{D \times D}$ \\

\addlinespace[0.5em]
\multicolumn{3}{@{}l}{\textit{Architecture mappings}} \\
\midrule
$\mathrm{xLSTM}[m\!:\!s](\cdot)$ & Hybrid xLSTM with $m$ mLSTM and $s$ sLSTM layers      & $\reals^{T \times D} \to \reals^{T \times D}$ \\
$\mathrm{Mamba\text{-}2}(\cdot)$  & Mamba-2 layer mapping                                  & $\reals^{T \times D} \to \reals^{T \times D}$ \\
$\mathrm{Gated DeltaNet}(\cdot)$             & Gated DeltaNet layer mapping                           & $\reals^{T \times D} \to \reals^{T \times D}$ \\

\addlinespace[0.5em]
\multicolumn{3}{@{}l}{\textit{Evaluation metrics}} \\
\midrule
$\mathrm{MASE}$           & Mean Absolute Scaled Error (time-series evaluation)          & scalar \\
$\mathrm{CRPS}$           & Continuous Ranked Probability Score (time-series evaluation) & scalar \\
$\mathrm{pass}@k$         & Code generation pass rate over $k \in \mathbb{N}$ samples    & scalar \\

\end{longtable}
\newpage

\section{Code-focused Language Model Pre-training Results}
\label{app:pretraining}

This appendix reports the full numerical results for the code-focused language-model pre-training experiments in Section~\ref{sec:pretraining}.
The main text focuses on the cross-family comparison between Gated DeltaNet, Mamba-2, and xLSTM\,$[7\!:\!1]$.
Here, we additionally report xLSTM\,$[1\!:\!0]$ and xLSTM\,$[11\!:\!1]$ to show how the linear-attention-to-recurrent-layer ratio affects performance within the xLSTM family.
All models are 400M-parameter inter-layer hybrids trained with the same recipe; only the sequence-operator configuration changes.

\paragraph{Code generation.}
Tables~\ref{tab:humaneval_64_20b_code}, \ref{tab:humaneval_64_100b_code}, and~\ref{tab:humaneval_64_20b_fineweb_code} report the full HumanEval pass@$k$ results.
Across all three training configurations, xLSTM\,$[7\!:\!1]$ is the best model at every reported pass@$k$.
The strongest non-xLSTM baseline changes with data and scale: Gated DeltaNet is second-best on both 20B-token and 100B code-only settings, while Mamba-2 is second-best on the mixed Nemotron-CC-Code-v1 + FineWeb-Edu corpus.
The additional xLSTM variants show that the xLSTM layer ratio matters: xLSTM\,$[1\!:\!0]$ and xLSTM\,$[11\!:\!1]$ are competitive in some cells, but neither matches the consistent HumanEval performance of xLSTM\,$[7\!:\!1]$. 

Enabling negative eigenvalues in the state-transition matrix~\citep{grazzi25unlocking} Gated DeltaNet\,$[-1\!:\!1]$ does not bring a substantial improvement for hybrid language model pretraining on code, as also observed by~\citet{merrill26olmo_hybrid}. 
However, in other tasks presented in Section~\ref{app:tsfm-implementation-details} and Sections~\ref{app:synthetic-implementation-details}, we observe that the specific state tracking capabilities enabled by the negative eigenvalues yield clear improvements.

\begin{table}[h]
    \centering
    \small
    \setlength{\tabcolsep}{6pt}
    \caption{HumanEval pass@$k$ ($k\in\{2,8,16,64\}$, \%) for 400M-parameter inter-layer hybrid variants pretrained on Nemotron-CC-Code-v1 for 20B tokens. Higher is better; the best result per column is shown in \textbf{bold}.}
    \begin{tabular}{lcccc}
    \toprule
    Model & pass@2 $\uparrow$ & pass@8 $\uparrow$ & pass@16 $\uparrow$ & pass@64 $\uparrow$ \\
    \midrule
    xLSTM\,$[1\!:\!0]$  & 1.42          & 3.67          & 5.34          & 8.53           \\
    xLSTM\,$[7\!:\!1]$  & \textbf{2.71} & \textbf{5.92} & \textbf{7.80} & \textbf{12.13} \\
    xLSTM\,$[11\!:\!1]$ & 2.60          & 5.16          & 6.82          & 10.20          \\
    Mamba-2             & 1.90          & 3.93          & 5.25          & 8.68           \\
    Gated DeltaNet      & 2.29          & 4.62          & 6.17          & 10.70          \\
    Gated DeltaNet\,$[-1,1]$ & 2.09 & 4.28 & 5.68 & 9.41\\
    \bottomrule
    \end{tabular}
    \label{tab:humaneval_64_20b_code}
\end{table}

\begin{table}[h]
    \centering
    \small
    \setlength{\tabcolsep}{6pt}
    \caption{HumanEval pass@$k$ ($k\in\{2,8,16,64\}$, \%) for 400M-parameter inter-layer hybrid variants pretrained on Nemotron-CC-Code-v1 for 100B tokens. Higher is better; the best result per column is shown in \textbf{bold}.}
    \begin{tabular}{lcccc}
    \toprule
    Model & pass@2 $\uparrow$ & pass@8 $\uparrow$ & pass@16 $\uparrow$ & pass@64 $\uparrow$ \\
    \midrule
    xLSTM\,$[1\!:\!0]$  & 3.21          & 6.66          & 8.54           & 13.09          \\
    xLSTM\,$[7\!:\!1]$  & \textbf{4.61} & \textbf{9.33} & \textbf{11.56} & \textbf{15.84} \\
    xLSTM\,$[11\!:\!1]$ & 4.04          & 7.42          & 9.41           & 13.14          \\
    Mamba-2             & 4.00          & 7.73          & 9.97           & 14.36          \\
    Gated DeltaNet      & 3.75          & 7.76          & 10.05           & 14.94          \\
    Gated DeltaNet\,$[-1,1]$ & 3.33 & 6.02 & 7.92 & 13.07\\
    \bottomrule
    \end{tabular}
    \label{tab:humaneval_64_100b_code}
\end{table}

\begin{table}[h]
    \centering
    \small
    \setlength{\tabcolsep}{6pt}
    \caption{HumanEval pass@$k$ ($k\in\{2,8,16,64\}$, \%) for 400M-parameter inter-layer hybrid variants pretrained on Nemotron-CC-Code-v1 + FineWeb-Edu for 20B tokens. Higher is better; the best result per column is shown in \textbf{bold}.}
    \begin{tabular}{lcccc}
    \toprule
    Model & pass@2 $\uparrow$ & pass@8 $\uparrow$ & pass@16 $\uparrow$ & pass@64 $\uparrow$ \\
    \midrule
    xLSTM\,$[1\!:\!0]$  & 0.88          & 1.96          & 3.01          & 6.40          \\
    xLSTM\,$[7\!:\!1]$  & \textbf{2.26} & \textbf{4.68} & \textbf{6.00} & \textbf{9.50} \\
    xLSTM\,$[11\!:\!1]$ & 1.81          & 3.29          & 4.20          & 7.17          \\
    Mamba-2             & 2.07          & 3.81          & 4.79          & 7.69          \\
    Gated DeltaNet      & 1.54          & 2.97          & 3.98          & 7.36          \\
    Gated DeltaNet\,$[-1,1]$ &1.26	&3.00	&4.20	&7.25          \\
    \bottomrule
    \end{tabular}
    \label{tab:humaneval_64_20b_fineweb_code}
\end{table}

\paragraph{Reasoning and commonsense.}
Tables~\ref{tab:benchmarks_pretraining_code_20B}, \ref{tab:benchmarks_pretraining_code_100B}, and~\ref{tab:benchmarks_pretraining} report the full reasoning and commonsense results.
Among the three cross-family backbones discussed in the main text, xLSTM\,$[7\!:\!1]$ has the best aggregate score in all three training configurations.
The margins are small compared with the HumanEval results, which supports the main-text conclusion that broad reasoning and commonsense evaluations are less sensitive to these backbone differences than code generation is.
The additional xLSTM-ratio ablations show a more mixed pattern: xLSTM\,$[11\!:\!1]$ gives the best aggregate score in the 20B-token  mixed corpus setting, while xLSTM\,$[7\!:\!1]$ gives the best aggregate score at pure code on both 20B and 100B tokens.

\begin{table*}[h]
    \centering
    \small
    \setlength{\tabcolsep}{3pt}
    \caption{Reasoning and commonsense benchmark results for 400M-parameter inter-layer hybrid variants pretrained on Nemotron-CC-Code-v1 for 20B tokens. Higher is better; the best result per column is shown in \textbf{bold}.}
    \resizebox{\textwidth}{!}{%
    \begin{tabular}{llcccccc}
    \toprule
    & & \multicolumn{5}{c}{Reasoning / Commonsense (\%)} & \\
    \cmidrule(lr){3-7}
    Training data & Model
    & HellaSwag $\uparrow$
    & PIQA $\uparrow$
    & ARC-Easy $\uparrow$
    & ARC-Challenge $\uparrow$
    & WinoGrande $\uparrow$
    & Avg. $\uparrow$
    \\
    \midrule
    \multirow{5}{*}{\shortstack[l]{Nemotron-CC-Code-v1\\(20B tokens)}}
    & xLSTM\,$[1\!:\!0]$  & 28.29          & 56.80          & 32.07          & 22.18          & \textbf{52.57} & 38.38          \\
    & xLSTM\,$[7\!:\!1]$  & 29.50          & \textbf{58.38}          & 33.88          & 21.16          & 50.91          & \textbf{38.76}          \\
    & xLSTM\,$[11\!:\!1]$ & \textbf{29.67} & 57.40 & 34.05 & 22.01 & 49.64          & 38.55 \\
    & Mamba-2             & 29.37          & 57.56          & \textbf{34.30}          & 21.42          & 50.75          & 38.68          \\
    & Gated DeltaNet      & 29.43          & 56.86          & 33.42          & 22.70          & 50.91          & 38.66          \\
    & Gated DeltaNet\,$[-1,1]$ & 29.29 & 56.69 & 33.08 & \textbf{23.38} & 49.80 & 38.45 \\
    \bottomrule
    \end{tabular}}
    \label{tab:benchmarks_pretraining_code_20B}
\end{table*}

\begin{table*}[h]
    \centering
    \small
    \setlength{\tabcolsep}{3pt}
    \caption{Reasoning and commonsense benchmark results for 400M-parameter inter-layer hybrid variants pretrained on Nemotron-CC-Code-v1 for 100B tokens. Higher is better; the best result per column is shown in \textbf{bold}.}
    \resizebox{\textwidth}{!}{%
    \begin{tabular}{llcccccc}
    \toprule
    & & \multicolumn{5}{c}{Reasoning / Commonsense (\%)} & \\
    \cmidrule(lr){3-7}
    Training data & Model
    & HellaSwag $\uparrow$
    & PIQA $\uparrow$
    & ARC-Easy $\uparrow$
    & ARC-Challenge $\uparrow$
    & WinoGrande $\uparrow$
    & Avg. $\uparrow$
    \\
    \midrule
    \multirow{5}{*}{\shortstack[l]{Nemotron-CC-Code-v1\\(100B tokens)}}
    & xLSTM\,$[1\!:\!0]$  & 30.13          & 58.38          & 34.68          & \textbf{23.29} & 49.49          & 39.19          \\
    & xLSTM\,$[7\!:\!1]$  & \textbf{31.38} & 57.56          & \textbf{36.32} & 22.18          & 52.09          & \textbf{39.91} \\
    & xLSTM\,$[11\!:\!1]$ & 30.11          & 57.94          & 35.23          & 21.76          & \textbf{52.17} & 39.44          \\
    & Mamba-2             & 30.06          & \textbf{58.98} & 35.48          & 21.67          & 50.91          & 39.42          \\
    & Gated DeltaNet      & 30.23          & 56.75          & 35.61          & 22.01          & 51.38          & 39.20          \\
    & Gated DeltaNet\,$[-1,1]$ & 29.64 & 58.76 & 34.68 & 21.59 & 48.93 & 38.72 \\

    \bottomrule
    \end{tabular}}
    \label{tab:benchmarks_pretraining_code_100B}
\end{table*}

\begin{table*}[h]
    \centering
    \small
    \setlength{\tabcolsep}{3pt}
    \caption{Reasoning and commonsense benchmark results for 400M-parameter inter-layer hybrid variants pretrained on Nemotron-CC-Code-v1 + FineWeb-Edu for 20B tokens. Higher is better; the best result per column is shown in \textbf{bold}.}
    \resizebox{\textwidth}{!}{%
    \begin{tabular}{llcccccc}
    \toprule
    & & \multicolumn{5}{c}{Reasoning / Commonsense (\%)} & \\
    \cmidrule(lr){3-7}
    Training data & Model
    & HellaSwag $\uparrow$
    & PIQA $\uparrow$
    & ARC-Easy $\uparrow$
    & ARC-Challenge $\uparrow$
    & WinoGrande $\uparrow$
    & Avg. $\uparrow$
    \\
    \midrule
    \multirow{5}{*}{\shortstack[l]{Nemotron-CC-Code-v1\\+ FineWeb-Edu\\(20B tokens)}}
    & xLSTM\,$[1\!:\!0]$  & 33.11          & 62.62          & 47.18          & 25.09          & 51.22          & 43.84          \\
    & xLSTM\,$[7\!:\!1]$  & \textbf{35.81} & \textbf{64.80} & \textbf{47.73}          & 25.91          & 49.72          & 44.79          \\
    & xLSTM\,$[11\!:\!1]$ & 35.36          & 64.25          & 46.97 & \textbf{26.88} & \textbf{52.64} & \textbf{45.22} \\
    & Mamba-2             & 34.98          & 64.15          & 46.97          & 25.60          & 49.64          & 44.27          \\
    & Gated DeltaNet      & 34.94          & 64.09          & 47.26          & 26.19          & 50.83          & 44.66          \\
    & Gated DeltaNet\,$[-1,1]$ & 35.00 & 63.71 & 46.00 & 26.19 & 51.62 & 44.50 \\
    \bottomrule
    \end{tabular}}
    \label{tab:benchmarks_pretraining}
\end{table*}

\section{Distillation Results}

\subsection{Distillation: Full Pass@\texorpdfstring{$k$}{k} on HumanEval}
\label{app:distill-code-passk}

Table~\ref{tab:distill-code-passk-app} reports the full pass@$k$ spread for $k\in\{1,2,8,16,32,64\}$ on HumanEval and HumanEval+ for the code distilled students discussed in Section~\ref{sec:distillation}.
All students follow the recipe described in Appendix~\ref{app:distillation-implementation-details}.
The xLSTM\,$[1\!:\!0]$ student remains strongest at every $k$ on both benchmarks.
Gated DeltaNet\,$[-1,1]$ improves over default Gated DeltaNet at every $k$ on HumanEval and HumanEval+, but does not close the gap to xLSTM\,$[1\!:\!0]$.

\begin{table}[h]
\centering
\small
\caption{Full pass@$k$ spread on HumanEval and HumanEval+ for the code distilled students.
Higher is better; the best student result per column is shown in \textbf{bold}.}
\label{tab:distill-code-passk-app}
\setlength{\tabcolsep}{5pt}
\begin{tabular}{llcccccc}
\toprule
Benchmark & Student & $k=1$ $\uparrow$ & $k=2$ $\uparrow$ & $k=8$ $\uparrow$ & $k=16$ $\uparrow$ & $k=32$ $\uparrow$ & $k=64$ $\uparrow$ \\
\midrule
\multirow{4}{*}{HumanEval}
 & Qwen3-4B-Instruct (teacher) & 0.914 & 0.927 & 0.944 & 0.950 & 0.956 & 0.963 \\
 & xLSTM\,$[1\!:\!0]$          & \textbf{0.831} & \textbf{0.882} & \textbf{0.940} & \textbf{0.952} & \textbf{0.956} & \textbf{0.957} \\
 & Gated DeltaNet              & 0.802 & 0.845 & 0.892 & 0.905 & 0.914 & 0.921 \\
 & Gated DeltaNet\,$[-1,1]$    & 0.813 & 0.854 & 0.897 & 0.912 & 0.922 & 0.927 \\
\midrule
\multirow{4}{*}{HumanEval+}
 & Qwen3-4B-Instruct (teacher) & 0.835 & 0.848 & 0.859 & 0.862 & 0.864 & 0.866 \\
 & xLSTM\,$[1\!:\!0]$          & \textbf{0.764} & \textbf{0.808} & \textbf{0.863} & \textbf{0.878} & \textbf{0.889} & \textbf{0.896} \\
 & Gated DeltaNet              & 0.739 & 0.780 & 0.832 & 0.846 & 0.856 & 0.866 \\
 & Gated DeltaNet\,$[-1,1]$    & 0.745 & 0.784 & 0.840 & 0.862 & 0.877 & 0.884 \\
\bottomrule
\end{tabular}
\end{table}

\subsection{Distillation Results on Math Data}
\label{app:distill-math}

Table~\ref{tab:distill-math-app} reports the math-distillation results for the same two student operators.
The xLSTM\,$[1\!:\!0]$ student leads on GSM8K and AIME 2024 pass@8, while Gated DeltaNet is slightly stronger on MATH-500.
Averaged across GSM8K, MATH-500, and AIME pass@8, xLSTM\,$[1\!:\!0]$ reaches 0.645 compared with 0.625 for Gated DeltaNet.
Table~\ref{tab:distill-aime-app} reports the full AIME 2024 pass@$k$ breakdown.

\begin{table}[h]
\centering
\small
\caption{Math distillation results for students distilled from Qwen3-4B-Instruct with the matched recipe of~\citet{hauzenberger26effective}.
GSM8K and MATH-500 report exact match~\citep{cobbe21gsm8k,hendrycks21math,lightman24lets}; AIME 2024 reports pass@8.
The Avg.\ column averages GSM8K, MATH-500, and AIME pass@8.
Higher is better; the best student result per column is shown in \textbf{bold}.}
\label{tab:distill-math-app}
\setlength{\tabcolsep}{6pt}
\begin{tabular}{lcccc}
\toprule
Student & GSM8K $\uparrow$ & MATH-500 $\uparrow$ & AIME pass@8 $\uparrow$ & Avg.\ $\uparrow$ \\
\midrule
Qwen3-4B-Instruct (teacher) & 0.939 & 0.854 & 0.367 & 0.720 \\
\midrule
xLSTM\,$[1\!:\!0]$          & \textbf{0.876} & 0.726 & \textbf{0.333} & \textbf{0.645} \\
Gated DeltaNet              & 0.842 & \textbf{0.732} & 0.300 & 0.625 \\
\bottomrule
\end{tabular}
\end{table}

\begin{table}[h]
\centering
\small
\caption{Full pass@$k$ breakdown on AIME 2024 for the two distilled students.
Higher is better; the best student result per column is shown in \textbf{bold}.}
\label{tab:distill-aime-app}
\setlength{\tabcolsep}{5pt}
\begin{tabular}{lcccccc}
\toprule
Student & $k=1$ $\uparrow$ & $k=2$ $\uparrow$ & $k=4$ $\uparrow$ & $k=8$ $\uparrow$ & $k=16$ $\uparrow$ & $k=32$ $\uparrow$ \\
\midrule
Qwen3-4B-Instruct (teacher) & 0.250 & 0.302 & 0.344 & 0.367 & 0.367 & 0.367 \\
\midrule
xLSTM\,$[1\!:\!0]$          & \textbf{0.055} & \textbf{0.105} & \textbf{0.190} & \textbf{0.333} & \textbf{0.333} & \textbf{0.333} \\
Gated DeltaNet              & 0.052 & 0.101 & 0.183 & 0.300 & 0.300 & 0.300 \\
\bottomrule
\end{tabular}
\end{table}

\section{Time-series Foundation-Model Pre-training Results}
\label{app:timeseries-results}

This appendix reports the full numerical results for the time-series foundation-model pre-training experiments in Section~\ref{sec:timeseries}.
All models use the same time-series pre-training protocol of~\citet{auer25tirex}; only the sequence mixer changes.
We evaluate zero-shot on GIFT-Eval~\citep{aksu24gift-eval} and report \ac{mase} and \ac{crps}, aggregated by geometric mean.
Lower values are better.

Table~\ref{tab:timeseries-results} shows the scaling comparison across five parameter scales.
We additionally also train Gated DeltaNet with the negative eigenvalues enabled in the state transition matrix, as described in \citet{grazzi25unlocking} and report the results in the table.
xLSTM\,$[3\!:\!1]$ gives the best result on both metrics from 1M to 40M parameters.
At 80M parameters, the architectures nearly converge: xLSTM,$[3\!:\!1]$ and Mamba-2 match on \ac{mase}, while Mamba-2 is best on \ac{crps}.
Notably, Gated DeltaNet with negative eigenvalues performs better than its non-negative counterpart, implying that time series foundation models benefit from the specific state tracking capacity enabled by this modification.

\begin{table}[htbp]
\caption{GIFT-Eval scores for time-series foundation models.
Models are evaluated zero-shot on GIFT-Eval, and results are aggregated by geometric mean.
Lower is better; the best result per scale and metric is shown in \textbf{bold}.}
\label{tab:timeseries-results}
\centering
\small
\resizebox{\textwidth}{!}{%
\begin{tabular}{lcccccccccc}
\toprule
 & \multicolumn{2}{c}{1M} & \multicolumn{2}{c}{4M} & \multicolumn{2}{c}{10M} & \multicolumn{2}{c}{40M} & \multicolumn{2}{c}{80M} \\
\cmidrule(lr){2-3} \cmidrule(lr){4-5} \cmidrule(lr){6-7} \cmidrule(lr){8-9} \cmidrule(lr){10-11}
Model
& \acs{mase} $\downarrow$
& \acs{crps} $\downarrow$
& \acs{mase} $\downarrow$
& \acs{crps} $\downarrow$
& \acs{mase} $\downarrow$
& \acs{crps} $\downarrow$
& \acs{mase} $\downarrow$
& \acs{crps} $\downarrow$
& \acs{mase} $\downarrow$
& \acs{crps} $\downarrow$ \\
\midrule
xLSTM\,$[3\!:\!1]$
& \textbf{0.768} & \textbf{0.527}
& \textbf{0.739} & \textbf{0.509}
& \textbf{0.733} & \textbf{0.508}
& \textbf{0.716} & \textbf{0.493}
& 0.715 & 0.490 \\
Mamba-2
& 0.795 & 0.545
& 0.784 & 0.533
& 0.767 & 0.525
& 0.721 & 0.496
& 0.715 & \textbf{0.485} \\
Gated DeltaNet
& 0.882 & 0.609
& 0.826 & 0.566
& 0.817 & 0.561
& 0.792 & 0.550
& 0.735 & 0.502 \\
 Gated DeltaNet\,$[-1,1]$
 & 0.792 & 0.541
 & 0.766 & 0.521
 & 0.764 & 0.517
 & 0.720 & 0.496
 & \textbf{0.714} & 0.489 \\
\bottomrule
\end{tabular}
}
\end{table}

\section{Synthetic Task Results}
\label{app:synthetic-results}

Table \ref{tab:synthetic-results} presents the results for all synthetic tasks across the three evaluated sequence lengths: 128 (training length), 512 and 2048.

\begin{table*}[h]
\centering
\caption{Length generalization of sequence mixers on synthetic counting and
state-tracking tasks. Models are trained at a sequence length of 128 and evaluated
at 128, 512, and 2048. Report accuracy (\%).}
\label{tab:synthetic-results}
\small
\setlength{\tabcolsep}{4pt}
\resizebox{\textwidth}{!}{

\begin{tabular}{l ccc ccc ccc | ccc ccc ccc}
& \multicolumn{9}{c}{\textbf{Counting}}
& \multicolumn{9}{c}{\textbf{State Tracking}} \\
\cmidrule(lr){2-10} \cmidrule(lr){11-19}
& \multicolumn{3}{c}{$A^n B^n$}
& \multicolumn{3}{c}{$A^n B^n C^n$}
& \multicolumn{3}{c}{Majority}
& \multicolumn{3}{c}{Parity}
& \multicolumn{3}{c}{Mod.\ Arith.}
& \multicolumn{3}{c}{$S_3$} \\
\cmidrule(lr){2-4} \cmidrule(lr){5-7} \cmidrule(lr){8-10}
\cmidrule(lr){11-13} \cmidrule(lr){14-16} \cmidrule(lr){17-19}
Model
& 128 & 512 & 2048
& 128 & 512 & 2048
& 128 & 512 & 2048
& 128 & 512 & 2048
& 128 & 512 & 2048
& 128 & 512 & 2048 \\
\midrule
xLSTM{[}1{:}0{]} & 1.000 & 0.963 & 0.892 & 1.000 & 0.984 & 0.932 & 1.000 & 0.982 & 0.763 & 0.013 & 0.001 & 0.003 & 0.314 & 0.292 & 0.283 & 0.088 & 0.022 & 0.003 \\
xLSTM{[}1{:}1{]} & 1.000 & 0.943 & 0.834 & 1.000 & 0.997 & 0.716 & 1.000 & 0.975 & 0.742 & 1.000 & 1.000 & 1.000 & 1.000 & 1.000 & 1.000 & 1.000 & 1.000 & 1.000 \\
Mamba-2           & 1.000 & 0.852 & 0.241 & 0.839 & 0.808 & 0.443 & 0.998 & 0.800 & 0.366 & 0.352 & 0.087 & 0.012 & 0.425 & 0.391 & 0.383 & 0.272 & 0.074 & 0.009 \\
Gated DeltaNet              & 1.000 & 0.993 & 0.820 & 0.981 & 0.805 & 0.322 & 0.991 & 0.717 & 0.268 & 0.060 & 0.014 & 0.003 & 0.361 & 0.340 & 0.328 & 0.141 & 0.036 & 0.004 \\
Gated DeltaNet[-1,1]        & 1.000 & 0.993 & 0.983 & 1.000 & 0.582 & 0.233 & 1.000 & 0.776 & 0.317 & 1.000 & 1.000 & 0.472 & 0.495 & 0.468 & 0.452 & 1.000 & 1.000 & 0.667 \\
\bottomrule
\end{tabular}

}
\end{table*}

\section{Experimental \& Implementation Details}
\label{app:expeirmental-details}

\subsection{Language Model Pretraining}
\label{app:lm-pretraining-implementation-details}

\begin{table}[t]
\centering
\caption{Hyperparameters for Language Model Pretraining at 400M scale.}
\label{tab:hyperparameters-lm-pretrain}
\begin{tabular}{ll}
\toprule
\textbf{Setting} & \textbf{Value} \\
\midrule
Hidden Size                           &  1024 \\
Num Layers                            &  24 \\
Positional Encoding                   &  NoPE \\
Num Heads (xLSTM/GDN)                       &  4 \\
Num Heads (Attention)                       &  16 \\
Num Heads (Mamba)                           &  64 \\
Conv Kernel Size                            &  4 \\
Context size                                &  8192 \\
Optimizer                                   &  AdamW \\
LR Scheduler                                &  Cosine + Warmup \\
Warmup \%                                   &  10\% \\
Weight decay                                &  0.1 \\
Learning rate                               & 3e-4 \\
Gradient Clipping                           &  1.0 \\
Effective Batch Size                        &  128 \\
Tokens per Step                        &  1M \\
Attention Layer idx                    & $ [6, 14, 22]$ \\
MLP Expansion Factor                   & 2.75 \\
Activation Function                    & SwiGLU \\
\bottomrule
\end{tabular}
\end{table}

\textbf{Pretraining Data.} We utilized the Nemotron-CC-Code-v1 and FineWeb Edu (100B subset) datasets, tokenized with GPT-NeoX tokenizer \citep{black22gpt-neox-20b}.

\textbf{Model Hyperparameters.}  Table \ref{tab:hyperparameters-lm-pretrain} reports hyperparameters of the pretrained language models and Table \ref{fig:lm-eval-setup} provides the evaluation details. For Gated DeltaNet, Mamba 2 and Attention we took the defaults settings from the \texttt{lm-engine} \citep{mishra24lm_engine} library and for xLSTM we took the setup of \citet{beck26xlstm} to achieve a matched 400m parameter count for all models.
All models are pre-norm architectures with RMSNorm.

\textbf{Training Setup. } All models were trained on 8xH100 GPUs using bfloat16 and PyTorch Distributed Data Parallel (DDP).

\begin{table}[ht]
\centering
\caption{Evaluation setup details for language model pre-training.}
\label{fig:lm-eval-setup}
\begin{tabular}{lrr}
\toprule
\textbf{Task} & \textbf{\# of shots} & \textbf{Generation Budget} \\
\midrule
\multicolumn{3}{l}{\textit{Language Understanding}} \\
\midrule
PIQA       & 0  & --   \\
ARC-e      & 0  & --   \\
ARC-c      & 25 & 100  \\
HellaSwag  & 10 & --   \\
Winogrande & 5  & --   \\
\midrule
\multicolumn{3}{l}{\textit{Language Generation}} \\
\midrule
HumanEval        & 0 & 1024 \\
\bottomrule
\end{tabular}
\label{tab:eval-tasks}
\end{table}

\subsection{Linearization via Distillation}
\label{app:distillation-implementation-details}

This appendix expands the distillation recipe of~\citet{hauzenberger26effective} used in Section~\ref{sec:distillation}.
The recipe replaces every softmax-attention block of a Transformer teacher with an intra-layer hybrid block that combines a linear-attention replacement with a sliding-window-attention path.
In our experiments, the linear-attention replacement is xLSTM\,$[1\!:\!0]$, default Gated DeltaNet, or, for code distillation, Gated DeltaNet\,$[-1,1]$.
The replacement inherits the teacher's $\vec{q}$, $\vec{k}$, and $\vec{v}$ projection weights at initialization.
Training proceeds in two stages: hidden-state alignment followed by sparse knowledge distillation.
We omit the optional expert-merging step of~\citet{hauzenberger26effective}.

\paragraph{Recipe compatibility.}
The recipe supports a candidate operator as a plug-in replacement when it exposes query, key, and value projections, admits a matrix-state formulation compatible with chunkwise-parallel kernels, and accepts the feature maps used by the scaffold.
xLSTM\,$[1\!:\!0]$, Gated DeltaNet, and Gated DeltaNet\,$[-1,1]$ satisfy these requirements.
xLSTM\,$[0\!:\!1]$ does not, because it is sequential and has no query-key-value analogue to initialize from the teacher.
Mamba-2 also requires a recipe extension: its input and forget gates are tied through a learned projection and scalar transition parameter, which do not have a direct teacher-attention analogue.
For this reason, Section~\ref{sec:distillation} isolates xLSTM\,$[1\!:\!0]$ and Gated DeltaNet variants as plug-in matrix-state operators.

\paragraph{Hybrid block.}
Each multi-head-attention block of the teacher is replaced by an intra-layer hybrid block whose output for token $t$ is
\begin{align}
    \hat{\vec{h}}_t
    =
    \vec{o}_t \odot \mathrm{LinAtt}(\vec{q}_t, \vec{k}_t, \vec{v}_t)
    +
    (\vec{1} - \vec{o}_t) \odot \mathrm{SWA}(\vec{q}_t, \vec{k}_t, \vec{v}_t),
\end{align}
where $\mathrm{LinAtt}$ is the linear-attention operator, $\mathrm{SWA}$ is sliding-window attention with window size 512 and four prepended sink tokens~\citep{xiao24efficient,beltagy20longformer}, and $\vec{o}_t \in (0, 1)^H$ is a per-head data-dependent sigmoid gate computed from the concatenation $[\vec{q}_t, \vec{k}_t, \vec{v}_t]$.
We apply rotary positional embeddings to $\vec{q}_t$ and $\vec{k}_t$.
The xLSTM\,$[1\!:\!0]$ branch uses head-wise softmax feature maps over the feature dimension and per-head scalar output gates; we keep the original xLSTM\,$[1\!:\!0]$ normalizer.

\paragraph{Stage I: hidden-state alignment.}
With teacher and student forward passes evaluated on the same input, we minimize the per-layer mean-squared error between teacher attention outputs $\vec{h}_t^{(\ell)}$ and student hybrid-block outputs $\hat{\vec{h}}_t^{(\ell)}$:
\begin{align}
    \mathcal{L}_{\mathrm{align}}
    =
    \sum_\ell \sum_t
    \big\|
        \vec{h}_t^{(\ell)} - \hat{\vec{h}}_t^{(\ell)}
    \big\|_2^2.
\end{align}
Only the newly introduced parameters, including feature maps and gates, are trainable in this stage.
Embeddings and feed-forward blocks remain frozen.

\paragraph{Stage II: sparse knowledge distillation.}
With all parameters unfrozen, we minimize a convex combination of next-token cross-entropy and a sparse top-$k$ KL divergence between the teacher and student output distributions:
\begin{align}
    \mathcal{L}_{\mathrm{KD}}
    =
    -\gamma \sum_t \log p_\theta(y_t \mid \vec{x}_{1:t})
    +
    \beta \sum_t
    \mathrm{KL}\!\left[
        p_T^{(k)}(\cdot \mid \vec{x}_{1:t})
        \,\big\|\,
        p_\theta^{(k)}(\cdot \mid \vec{x}_{1:t})
    \right],
\end{align}
where $p_T^{(k)}$ and $p_\theta^{(k)}$ are the top-$k$ truncations of the teacher and student distributions.
We set $\gamma = 0.9$, $\beta = 0.1$, and $k = 256$.
Sparsifying the teacher distribution allows precomputing teacher targets, so the teacher does not need to run during Stage II.

\paragraph{Hyperparameters.}
Sequence length is 4{,}096 throughout.
Stage II trains for 10{,}000 optimization steps in each domain.
Code distillation uses Nemotron-Pretraining-Code-v2~\citep{nvidia25nemotron3nano}.
Math distillation uses Nemotron-Math-v2~\citep{du25nemotron-math}.
Optimizer, learning-rate schedule, and remaining hyperparameters follow~\citet{hauzenberger26effective} unchanged.

\paragraph{Gated DeltaNet variants.}
The Gated DeltaNet branch follows the recipe defaults: rotary-positioned queries and keys, SiLU feature maps applied to $\vec{q}$, $\vec{k}$, and $\vec{v}$, $qk$ $L^2$-normalization inside the kernel, recurrent-gate clamping at $-3$, document-boundary resets with reset value $-25$, and an unmerged background gating head.
The recurrent branch uses the chunk kernel during training and inference.
For code distillation, Gated DeltaNet\,$[-1,1]$ keeps the same scaffold, data, initialization, and optimization recipe, but uses the negative-eigenvalue parameterization of~\citet{grazzi25unlocking}.
We do not enable this variant in the math-distillation runs, because the corresponding xLSTM\,$[m\!:\!s]$ extension is outside the scope of the matched plug-in comparison.

\subsection{Pretraining Time Series Foundation Models}
\label{app:tsfm-implementation-details}

\textbf{Pretraining Data. } 
The time series foundation models were pretrained on a corpus of $\sim 47.5$M timeseries, based on \citet{auer25tirex}.
This consists of $\sim 30$M series from the Chronos pretraining corpus introduced in \citet{ansari24chronos}, $\sim 2.5$M series from the GIFT-Eval Pretraining corpus introduced in \citet{aksu24gift-eval}, and $\sim 15$M series synthetically generated using KernelSynth \citep{ansari24chronos}.
The data corpus has been cleaned for zero overlap with the GIFT-Eval training corpus, hence all evaluations are zero-shot.

\textbf{Training Setup. } All time series models were trained on 4x NVIDIA A100 GPUs using bfloat16 and PyTorch Distributed Data Parallel (DDP). All model hyperparameters are reported in Tables \ref{tab:hyperparameters-tsfm-pretrain} and \ref{tab:tsfm-param-sweep}

\begin{table}[h]
\centering
\small
\setlength{\tabcolsep}{4pt}
\caption{\textbf{Architecture hyperparameters for the five parameter scales used throughout \ac{tsfm} experiments.} Width (hidden dimension), depth (number of layers), input/output projection (``linear'' at 1M, residual-MLP with the listed hidden dimension otherwise), MLP expansion factor and number of heads. All backbones at a given scale share these settings.}
\label{tab:tsfm-param-sweep}
\begin{tabular}{lrrlrr}
\toprule
\textbf{Scale} & \textbf{Width} & \textbf{Depth} & \textbf{In/Out proj.} & \textbf{MLP exp.} & \textbf{Num Heads} \\
\midrule
1M  & 128 & 6  & linear    & 2.0 & 4 \\
4M  & 256 & 6  & MLP, 1024 & 2.0  & 4\\
10M & 384 & 6  & MLP, 2048 & 2.0  & 3\\
40M & 512 & 12 & MLP, 2048 & 2.75  & 4\\
80M & 768 & 12 & MLP, 2048 & 2.75  & 6\\
\bottomrule
\end{tabular}
\label{tab:arch-hparams}
\end{table}

\begin{table}[t]
\centering
\caption{Shared Training Parameters for Time Series Foundation Models across scale}
\label{tab:hyperparameters-tsfm-pretrain}
\begin{tabular}{ll}
\toprule
\textbf{Setting} & \textbf{Value} \\
\midrule
Positional Encoding                   &  NoPE \\
Conv Kernel Size                            &  4 \\
Context size                                &  8192 \\
Patch Size                                  &  32 \\
Gradient Clipping                           &  1.0 \\
Optimizer                                   &  AdamW \\
LR Scheduler                                &  Cosine + Warmup \\
Warmup \%                                   &  3 \% \\
Weight decay                                &  0.1 \\
Learning rate                               & 1.2e-3 \\
Effective Batch Size                        &  256 \\
Num Steps                                   &  500,000 \\
Activation Function                    & SwiGLU \\
\bottomrule
\end{tabular}
\end{table}

\subsection{Synthetic Counting and State-tracking Experiments}
\label{app:synthetic-implementation-details}

\textbf{Tasks}. We evaluate sequence mixers on six synthetic formal-language tasks split into two families. The counting family contains $A^nB^n$ (balanced two-symbol strings), $A^nB^nC^n$ (balanced three-symbol strings), and Majority (predict the majority symbol over an input sequence). The state-tracking family contains Parity (predict the running XOR of a binary sequence), Modular Arithmetic over $\mathbb{Z}_5$ (running sum modulo 5), and word-problem evaluation in the symmetric group $S_3$ (composition of permutations on three elements). 

\textbf{Data}. For each task, training samples are drawn at a sequence length of 128. Evaluation samples are drawn at three lengths: 128 (in-distribution), 512 (4× extrapolation), and 2048 (16× extrapolation). At each evaluation length, we sample a held-out test set disjoint from training. The model is required to output the task target at the final position of the sequence.

\textbf{Training Setup}. All models are trained from scratch at a sequence length of 128. Models are trained on a single NVIDIA H100 GPU using bfloat16. We used two block structures without MLP layers. We report all training hyperparameters in Table \ref{tab:synthetic_train_details}.

\textbf{Evaluation}. We report token-level accuracy at the target position, averaged over the held-out evaluation set at each length. We run each experiment over 5 seeds and report the maximum of those runs.

\begin{table}[h]
\centering
\caption{Training hyperparameters for synthetic task experiments.}
\label{tab:synthetic_train_details}
\begin{tabular}{ll}
\toprule
Setting & Value \\
\midrule
Hidden Size & 128 \\
Num Layers & 2 \\
Num Heads & 4 \\
Positional Encoding & NoPE \\
Training Sequence Length & up to 128 \\
Evaluation Sequence Lengths & 128, 512, 2048 \\
Optimizer & AdamW \\
LR Scheduler & Cosine + Warmup \\
Warmup \% & 3 \% \\
Weight Decay & \{1e-1, 1e-3, 1e-4\} \\
Learning Rate & \{1e-3, 3e-4, 1e-4\} \\
Gradient Clipping & 1.0 \\
Batch Size & 256 \\
Num Steps & 50,000 \\
Num Seeds & 5 \\
\bottomrule
\end{tabular}
\end{table}

\section{Related Linearization Work}
\label{app:distill-related-work}

Existing linearization work converts Transformer language models into subquadratic students, but typically fixes the target operator family.
LoLCATs~\citep{zhang25lolcats} replaces softmax attention with an intra-layer hybrid of linear and sliding-window attention, fitting the linearization at the layer level rather than comparing candidate sequence mixers.
Liger~\citep{lan25liger} linearizes language models into gated recurrent structures through gating-only modifications.
RADLADS~\citep{goldstein25radlads} distills RWKV-6/7 backbones with an RWKV-specific pipeline.
Llamba~\citep{bick25llamba} converts attention layers to Mamba-2 state-space mixers, and MOHAWK~\citep{bick24mohawk} aligns Transformer hidden states to subquadratic state-space students.
Mamba-in-Llama~\citep{wang24mamba} interleaves Mamba layers with surviving attention layers in a hybrid student.
These works establish that Transformer linearization is feasible, but they do not compare xLSTM\,$[1\!:\!0]$ and Gated DeltaNet under the same teacher, data, scaffold, and optimization recipe.
Section~\ref{sec:distillation} isolates this matched plug-in comparison.

\end{document}